%% file: main.tex
\pdfoutput=1
\documentclass[11pt]{article}
\usepackage{amsmath,amssymb,booktabs}
\usepackage[]{acl}
\usepackage{svg}
\usepackage{enumitem}
\usepackage{hyperref}
\usepackage[hyphenbreaks]{breakurl}
\usepackage{breakurl}
\usepackage{subfigure}
\usepackage{booktabs} % for professional tables
\usepackage{multirow}
\usepackage{times}

\usepackage{latexsym}
\newif\ifshowcomment
    \showcommenttrue
\ifshowcomment
    \newcommand{\yang}[1]{\textcolor{blue}{[yang: #1]}}

\else
    \newcommand{\todo}[1]{}

    \newcommand{\yang}[1]{}

\fi

\usepackage{hyperref}

\usepackage[T1]{fontenc}
\usepackage[utf8]{inputenc}

\usepackage{microtype}

\title{Making Pre-trained Language Models both 
\\ Task-solvers and Self-calibrators}

\author{Yangyi Chen \\
  UIUC \\
  \And
  Xingyao Wang \\
  UIUC \\
  \And
  Heng Ji\\
  UIUC
  \AND
{\tt yangyic3@illinois.edu}
  }

\begin{document}
\maketitle
\begin{abstract}
% \looseness=-1
Pre-trained language models (PLMs) serve as backbones for various real-world systems. For high-stake applications, it's equally essential to have reasonable confidence estimations in predictions. While the vanilla confidence scores of PLMs can already be effectively utilized, PLMs consistently become overconfident in their wrong predictions, which is not desirable in practice. Previous work shows that introducing an extra calibration task can mitigate this issue. The basic idea involves acquiring additional data to train models in predicting the confidence of their initial predictions. However, it only demonstrates the feasibility of this kind of method, assuming that there are abundant extra available samples for the introduced calibration task. In this work, we consider the practical scenario that we need to effectively utilize training samples to make PLMs both task-solvers and self-calibrators. Three challenges are presented, including limited training samples, data imbalance, and distribution shifts. We first conduct pilot experiments to quantify various decisive factors in the calibration task. Based on the empirical analysis results, we propose a training algorithm LM-TOAST to tackle the challenges. Experimental results show that LM-TOAST can effectively utilize the training data to make PLMs have reasonable confidence estimations while maintaining the original task performance. Further, we consider three downstream applications, namely selective classification, adversarial defense, and model cascading, to show the practical usefulness of LM-TOAST. The code will be made public at \url{https://github.com/Yangyi-Chen/LM-TOAST}.

% The code will be released.
\end{abstract}

\input{intro}

\input{background}

\input{pilot}

\input{method}

\input{exp}
\input{application}

\input{related}

\section{Conclusion}
% \looseness=-1
We present the task-agnostic LM-TOAST to make PLMs have reasonable confidence estimations while maintaining the original task performance.
% Experimental results show that LM-TOAST makes PLMs better self-calibrators, evidenced by the better discrimination in confidence scores of correct and wrong predictions. 
We also show that its good self-calibration can be transferred to downstream applications.

% 

% 
% By analyzing three decisive factors in the calibration task training, we propose a method named LM-TOAST.
% 

% \section{Individual Contributions}
% \label{sec:individual}
% Yangyi makes the research proposal and initializes the research plan.
% Xiaofei participates in the discussion. 
% For experiments, Xiaofei implements some baseline methods, including temperature scaling and label smoothing.
% Yangyi implements the proposed method in this paper and conducts most experiments. 
% For paper writing, Yangyi writes most of the paper and Xiaofei helps to organize the results.
% % 
% Also, Xingyao Wang gives valuable suggestions on this project and helps to improve the paper writing and organization. 

\input{limitation}

% \newpage
% \quad
% \newpage

\bibliography{main}
\bibliographystyle{acl_natbib}
\newpage
\quad
\newpage
\appendix
\input{appendix}

\end{document}

%% file: intro.tex
\section{Introduction}

\begin{figure}[h]
    \centering
    \includegraphics[width=0.7\linewidth]{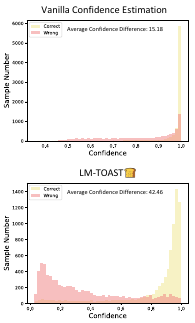}
    % \vspace{-20pt}
    \caption{\looseness=-1 The confidence distributions of correct and wrong predictions on the Amazon review sentiment analysis dataset. ``Average Confidence Difference'' measures the difference in average confidence between correct and wrong predictions. }
    \vspace{-15pt}
    \label{fig:hello}
\end{figure}

% \looseness=-1
We have witnessed the great success of pre-trained language models (PLMs) over the past few years in various tasks~\citep{DBLP:conf/nips/WangPNSMHLB19, DBLP:conf/iclr/WangSMHLB19}.
Nowadays, real-world natural language processing (NLP) systems are mostly built upon PLMs to effectively utilize their strong capacities~\citep{DBLP:journals/corr/abs-2108-07258, DBLP:journals/aiopen/HanZDGLHQYZZHHJ21}. 

\looseness=-1
Beyond the performance evaluation, an essential requirement in high-stake applications for PLMs is to assign reasonable confidence to their predictions. 
This can enable the decision-makers to better handle the low-confident predictions, e.g., directly abstain to give predictions or transfer the input to human experts. 
The original predictive probability of PLMs can be effectively utilized for ranking predictions. 
This simple strategy can reasonably give relatively higher confidence to correct predictions compared to the wrong ones~\citep{DBLP:conf/acl/HendrycksLWDKS20}. 
% 

% as their confidence estimations.
% 
% 
% This simple strategy can reasonably rank predictions based on models' confidence. 

% This simple strategy can well 
% 

\looseness=-1
However, relying on the vanilla confidence scores cannot well distinguish between correct and wrong predictions.
PLMs consistently assign high confidence in their predictions, no matter correct or not~\citep{DBLP:journals/corr/abs-2211-00151}.
This results in a large number of wrong predictions distributed in the high-confident zone (see Figure~\ref{fig:hello}).
The direct undesirable consequence is the false acceptance of wrong but high-confident predictions.
Besides, previous work avoids the issue of selecting a concrete confidence threshold by using hyper-parameter-free metrics (e.g., AUROC) in relevant tasks (e.g., selective classification).
% various practical tasks rely on models' confidence to help practitioners make better decisions (e.g., selective classification). 
% 
% 
But in practice, the small gap between confidence in correct and wrong predictions may cause large performance variance due to the manually chosen threshold. 

% But in practice, a manually chosen threshold is incompatible with the fine-grained confidence difference, which may result in a significant performance drop due to some small variance in the chosen value. \xw{But what does this means? a slightly change of the probability threshold will result in significant performance drop? }

% a confidence threshold should be chosen in practice. 
% Previous work avoids this issue 

% Besides, for various tasks (e.g., selective classification), 
% previous work aleaves out this problem by adopting hyper-parameter-free metrics (e.g., AUROC). 
% 
% Thus, there is no need to choose a concrete confidence threshold. \xw{This part is a bit confusing?}
% 

% more robust interval for the confidence level chosen is desirable. 
% 

Existing work shows that an extra calibration task can be taken as a remedy~\citep{DBLP:journals/corr/abs-2211-00151, lin2022teaching}. 
The calibration task uses extra samples to train models to have reasonable confidence estimations. 
However, previous work considers ideal situations to demonstrate the feasibility, assuming access to a large number of unused labeled samples, typically from the validation set.
In practice, the samples in the validation dataset may be too small to guarantee good calibration performance.
Besides, relying on the validation samples for the calibration task training causes data leakage, which may result in unreliable performance estimation when adopting the validation dataset to choose hyper-parameters. 
In practice, we need to effectively utilize training samples for both the original and the calibration tasks training. 
Three challenges are presented: 
\begin{itemize} [topsep=1pt, partopsep=1pt, leftmargin=12pt, itemsep=-2pt]
\item \textbf{Limited training samples}: How to effectively utilize the training samples to increase the calibration task performance while maintaining the original task performance? 
\item \textbf{Data imbalance}: Given PLMs' high performance, the positive cases (correctly classified samples) significantly dominate the calibration training set, causing the data imbalance issue. 
\item \textbf{Distribution shifts}: When deployed, PLMs are also expected to exhibit out-of-distribution (OOD) robustness, assigning reasonable confidence scores to OOD samples.
\end{itemize}
% 

% \looseness=
In this work, we motivate to make PLMs both task-solvers and self-calibrators in practical settings. 
% 
% We situate our method in practice and consider the practical setting when only limited training samples are given. 
% 
We first conduct pilot experiments to quantify various decisive factors in the calibration task, including the number of training samples, the data imbalance ratio, and input ``features'' for the calibration task. 
Based on the empirical analysis, we propose a training algorithm LM-TOAST to tackle the challenges.
We employ K-fold cross-annotation to generate the training data for the calibration task. 
Then we employ data down-sampling, adversarial data augmentation, and consistent training to tackle the challenges of data imbalance and distribution shifts.
Note that LM-TOAST can be applied to all classification tasks to improve confidence estimations.
Experimental results show that LM-TOAST can increase the discrimination between correct and wrong predictions, evidenced by the fine-grained order ranked by confidence and a sharp difference in average confidence.
% 

% \looseness=-1
Further, we show that the good self-calibration performance of LM-TOAST can be transferred to downstream applications.
We consider three downstream tasks, including selective classification~\citep{geifman2017selective}, adversarial defense~\citep{DBLP:journals/tist/ZhangSAL20}, and model cascading~\citep{varshney2022model}.
Experimental results demonstrate the practical significance of LM-TOAST in these applications.

% (1) Selective classification: In high-stake applications, models may reject to give low-confident predictions and the corresponding samples may be transferred to human experts to maintain performance; 
% % 
% (2) Adversarial defense: Textual adversarial samples may introduce perturbations to confuse the predictions. The low confidence may serve as a useful signal to detect malicious inputs;
% % 
% (3) Model cascading: Real-world NLP systems may retain a large model pool, consisting of various sizes of PLMs. 
% Small PLMs are first adopted for predictions.
% While for inputs with lower confidence scores, larger PLMs can be adopted for solutions, ensuring the performance and efficiency of the whole system.
% 

%% file: background.tex
\section{Background}
\subsection{Task Formalization}
\label{sec:task_formalization}
In standard classification training, a model $\mathcal{F}:\mathbb{X}\rightarrow \mathbb{Y}$ for the main task is trained on a given dataset $\mathbb{D}=\{(x_i,y_i)_{i=1}^{N}\}$ to minimize the pre-defined classification loss.
For the introduced calibration task, a new calibration dataset $\mathbb{D}^*=\{(x_i,y_i^*,c_i)_{i=1}^{M}\}$ is generated from $\mathbb{D}$, where $x_i$ is the original sample in $\mathbb{D}$, $y_i^*$ is model's prediction, and $c_i$ is the corresponding confidence score. 
The calibration task aims to predict the models' confidence using the sample and the original prediction. 
The generation process of $\mathbb{D}^*$ is one essential part of the calibration task.
\citet{lin2022teaching} propose to deem accuracy on a batch of samples as the confidence $c_i$ for samples in this batch. 
In this work, we simplify this assumption and directly treat $c_i$ as a binary value, indicating whether the prediction $y_i^*$ is correct or not. 
% 
% 
% ~\citep{liu}
\input{tabs/template_example}

\input{figs/quantify_1}

\looseness=-1
Once $\mathbb{D}^*$ is generated, one can fit an extra model $\mathcal{F^*}$ separately or conduct multi-task training using the original model $\mathcal{F}$ to learn the calibration task\footnote{Due to the unified modeling approach proposed by~\citet{DBLP:journals/jmlr/RaffelSRLNMZLL20}, $\mathcal{F}$ can be easily utilized as a mapping function $\mathcal{F}:(\mathbb{X},\mathbb{Y}) \rightarrow \mathbb{C}$.}.
In this work, we adopt the latter paradigm since no obvious performance difference is observed in previous work~\citep{DBLP:journals/corr/abs-2211-00151}. 
Specifically, we adopt the unified text-to-text paradigm and use T5 as the backbone model in this paper~\citep{DBLP:journals/jmlr/RaffelSRLNMZLL20}. 
We use two sets of templates and verbalizers for the main task and the calibration task respectively~\citep{liu2021pre}.
See Table~\ref{tab:template_example} for an example used in the sentiment analysis task.
Other templates and verbalizers selected are listed in Appendix~\ref{sec:appendix:template}. 
The probability of the ``True'' class in the calibration task is deemed as PLMs' confidence in their predictions. 
In testing, the original test set is used for evaluating both the original and the calibration tasks. 

\input{tabs/dataset}

\subsection{Evaluation Setting}
\looseness=-1
\paragraph{Evaluation metric.}
% Given the confidence score PLMs assign to each correct or wrong prediction, 
We adopt two evaluation metrics to characterize whether PLMs assign reasonable confidence to testing samples that consist of correct and wrong predictions:
(1) \textbf{AUROC} (Area Under the Receiver Operating Characteristic curve), which doesn't require manually picking a threshold value~\citep{DBLP:conf/icml/DavisG06}. A better AUROC score indicates that correct predictions have relatively higher confidence scores than wrong ones;
(2) \textbf{$\Delta$Conf}, which directly measures the average confidence difference between correct and wrong predictions. 
A higher $\Delta$Conf score indicates a better distinction between correct and wrong predictions from the confidence scores.
Note we don't use ECE~\citep{naeini2015obtaining} since we mostly consider relative confidence scores in this work. See \citet{fisch2022uncertainty} for detailed elaborations.

\paragraph{Evaluation dataset.}
For all experiments in this paper, we evaluate in both in-distribution (ID) and out-of-distribution (OOD) settings. 
We consider three classic tasks, including sentiment analysis, hate-speech detection, and natural language inference. 
We follow the same dataset chosen in \citet{DBLP:journals/corr/abs-2211-00151} (see Table~\ref{tab:dataset}).
% 
% The datasets chosen are listed in Table~\ref{tab:dataset}, and 
The detailed descriptions and references are in Appendix~\ref{appendix:dataset}. 
% The templates and verbalizers are in Appendix. 
% 

%% file: tabs/template_example.tex
\begin{table}[]
\centering
\resizebox{0.95\linewidth}{!}{
\begin{tabular}{lll}
\toprule
Task        & Template                                                                                                                                                                                                                              & Verbalizer               \\ \midrule
Main        & It was \textless{}mask\textgreater{},  \textless{}input\_sentence\textgreater{}                                                                                                                                                       & {[}bad, good, neutral{]} \\ \midrule
Calibration & \begin{tabular}[c]{@{}l@{}}Sentence: \textless{}input\_sentence\textgreater{},  The predicted sentiment is:\\  \textless{}prediction\textgreater{}. Is the prediction True or False? It's \textless{}mask\textgreater{}.\end{tabular} & {[}False, True{]}        \\ \bottomrule
\end{tabular}
}
% \vspace{-5pt}
\caption{Example of templates and verbalizers used in the sentiment analysis task. \textless{}input\_sentence\textgreater{} denotes the original sample. \textless{}prediction\textgreater{} denotes the original prediction. Others are shown in Appendix~\ref{sec:appendix:template}.}
% \vspace{-15pt}
\label{tab:template_example}
\end{table}

%% file: figs/quantify_1.tex
\begin{figure*}
  \centering
  \includegraphics[width=0.32\textwidth]{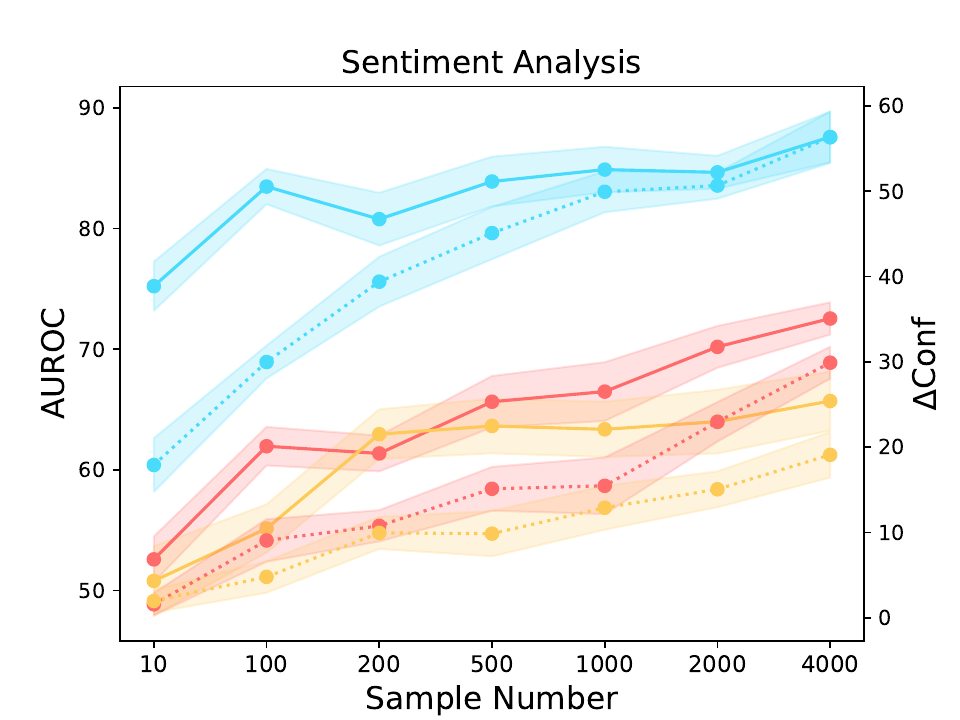}
%   \hspace{1in}
  \includegraphics[width=0.32\textwidth]{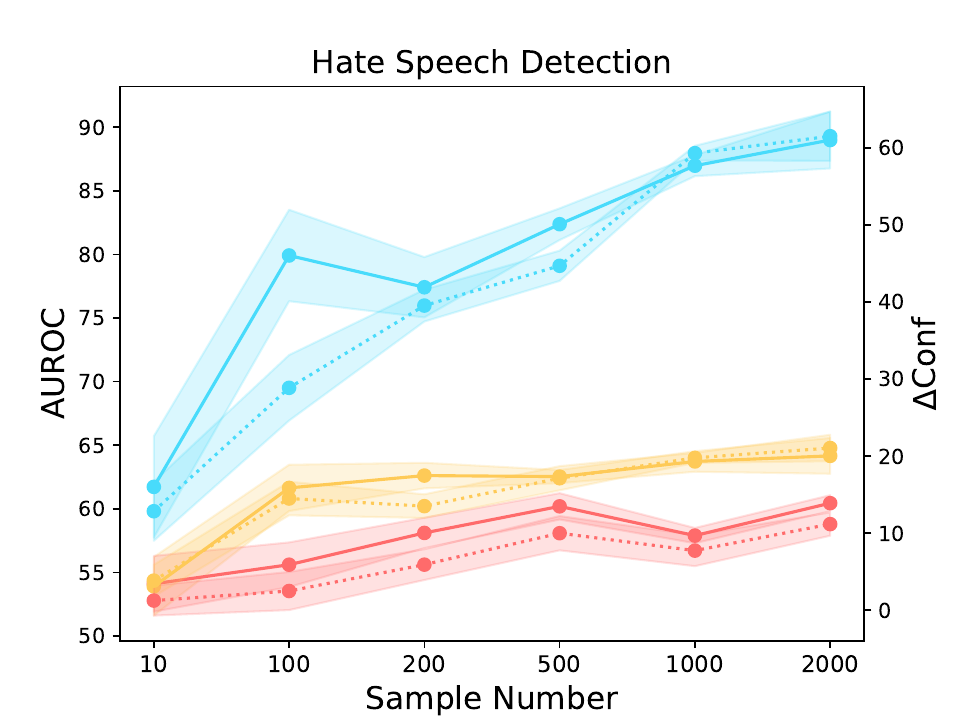}
%   \hspace{1in}
  \includegraphics[width=0.32\textwidth]{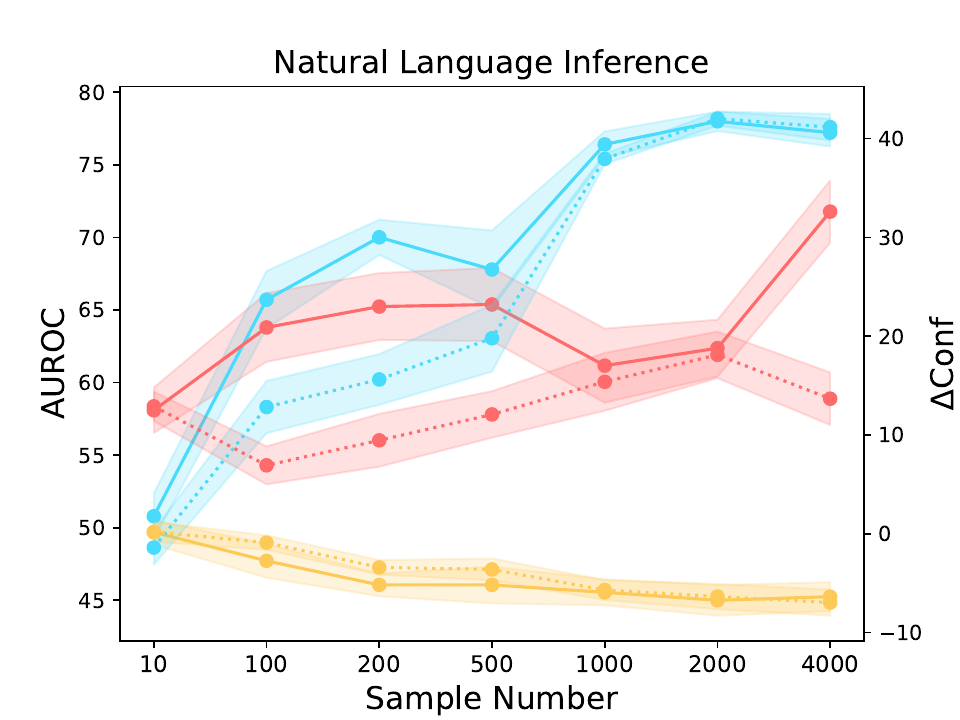}
   \hspace{1in}
   \includegraphics[width=0.95\textwidth]{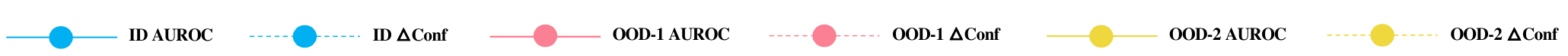}
  
  \caption{Quantify the influence of available training samples in the calibration task. The evaluation datasets (ID, OOD-1, OOD-2) are listed in Table~\ref{tab:dataset}. Increasing the dataset size of the calibration task continually brings benefits.}
  \vspace{-10pt}
  \label{fig:quantify_1}
\end{figure*}

%% file: tabs/dataset.tex
\begin{table}[]
\centering
\resizebox{0.9\linewidth}{!}{
\begin{tabular}{l|c|cc}
\toprule
Task                       & ID             & OOD-1  & OOD-2 \\ \midrule
Sentiment Analysis         & Amazon         & SST-5        & SemEval  \\
Hate-speech Detection      & Civil & Hate Speech  & Implicit     \\
Natural Language Inference & MNLI           & HANS         & ANLI     \\ \bottomrule
\end{tabular}
}
% \vspace{-5pt}
\caption{The chosen evaluation datasets for each task.}
\vspace{-15pt}
\label{tab:dataset}
\end{table}

%% file: pilot.tex
% \vspace{-10pt}

% \vspace{-20pt}

% \vspace{-10pt}

\input{figs/quantify_2}

\section{Pilot Experiments and Analysis}
We conduct pilot experiments to quantify the influence of several decisive factors in the calibration task, which can help for a better design of various components in the training algorithm. 
Specifically, we consider the number of training samples, dataset imbalance, and input ``features''. 
The concrete experimental settings are described in Appendix~\ref{sec:appendix:summary}.
% 
% \vspace{-18pt}

\subsection{Number of Training Samples}
\label{sec:num_train}
We quantify the influence of available training samples in the calibration task (see Figure~\ref{fig:quantify_1}). 
We observe overall consistent trends in three different tasks across eight datasets. 
The results show the continued benefits of increasing the dataset size for the calibration task considering the AUROC and $\Delta$Conf scores. 
Surprisingly, the performance in OOD datasets can be improved by introducing more calibration training samples in ID datasets. 
This is different from the common belief that learning more in-domain knowledge may hurt the OOD robustness due to the reliance on spurious correlations~\citep{DBLP:conf/icml/RadfordKHRGASAM21}.
However, we note that there is an unnatural trend in the natural language inference task when ANLI is adopted as the OOD evaluation dataset. 
The reason may be partially attributed to the unique construction process of ANLI based on human-in-the-loop attacks.
% 
% Thus, the constructed samples are similar to ID samples but with misleading features \xw{maybe elaborate more clearly on what type of misleading feature? or we can just remove this sentence I guess}, which may result in the inverse trend in the calibration task. 
% 

\subsection{Data Imbalance}
We vary the class ratios in the calibration training set to quantify the influence of data imbalance in the calibration task (see Figure~\ref{fig:quantify_2}). 
Note that there are two classes in the calibration training set, where the positive (negative) case indicates that the model's original prediction is correct (wrong). 
The class ratio is defined as the fraction of negative-class samples in the whole dataset. 
We consistently observe inverted V-shapes considering all evaluation settings. 
Thus, we draw the conclusion that given a calibration dataset with a fixed number of samples, an exact balanced distribution of the two classes will mostly benefit the calibration task.

\looseness=-1
Further, we consider a more practical question faced in our algorithm: Given limited training samples in one class, what is the influence of continuing to increase the training samples in the other class? 
We conduct two rounds of experiments. 
For each round, we fix the samples in one class and continue to increase the samples in the other class. 
The results are shown in Figure~\ref{fig:quantify_further}. 
We observe roughly V-shapes in both two rounds. 
Also, we carefully observe the two ends of the V shapes and find that these two dataset scaling processes can hardly bring a positive effect on PLMs' calibration performance. 
Thus, given limited available training samples, the optimal strategy is to keep the
dataset with an exact balanced distribution of two classes even if we have extra data because we cannot precisely predict how many samples we should add to one single class to improve the calibration performance.

% Namely, the optimal strategy is to keep the dataset with an exact balanced distribution of two classes even if we have extra data. 

% \vspace{-10pt}

\subsection{Input ``Features''}
% \looseness=-1
Recall from Sec.~\ref{sec:task_formalization} that two ``features'' exist in each calibration training sample $(x_i,y_i^*,c_i)$, including the original sample $x_i$ and the model's original prediction $y_i^*$. 
We ablate the influence of these two ``features'' (see Table~\ref{tab:quantify_3}). 
We observe the dominant effect of information extracted from the original sample $x_i$.
While still lagging behind the calibration performance when using all features, only relying on the original sample for the prediction can achieve descent performance in most cases. 

The experimental results can further inform us of the essence of the calibration task. 
Given the descent performance when only using the original samples as the input features, PLMs mostly are performing the task of determining the difficulty of each sample, where hard (easy) samples will be assigned low (high) confidence scores. 
The potential major function PLMs learn in the calibration task training may be  inducing which kinds of features in the texts they cannot handle well.
In this work, we motivate to exploit the calibration methods, and further exploration and utilization of the inner mechanism are left for future work. 

\input{tabs/quantify_3}

% \subsection{Summary} 
% \looseness=-1

%% file: figs/quantify_2.tex
\begin{figure*}
  \centering
  \includegraphics[width=0.32\textwidth]{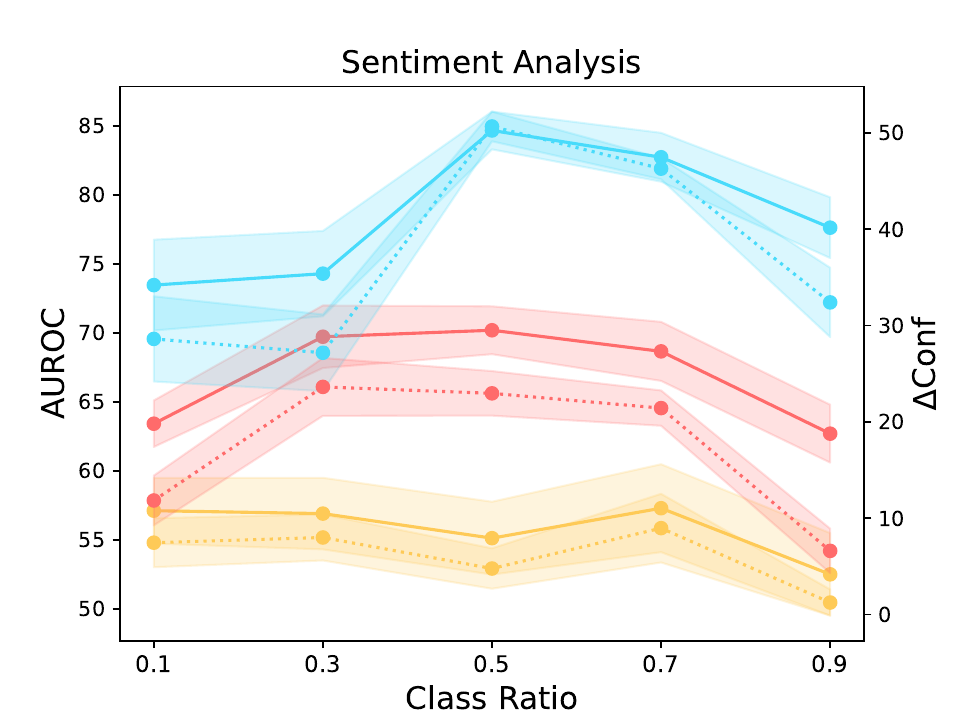}
%   \hspace{1in}
  \includegraphics[width=0.32\textwidth]{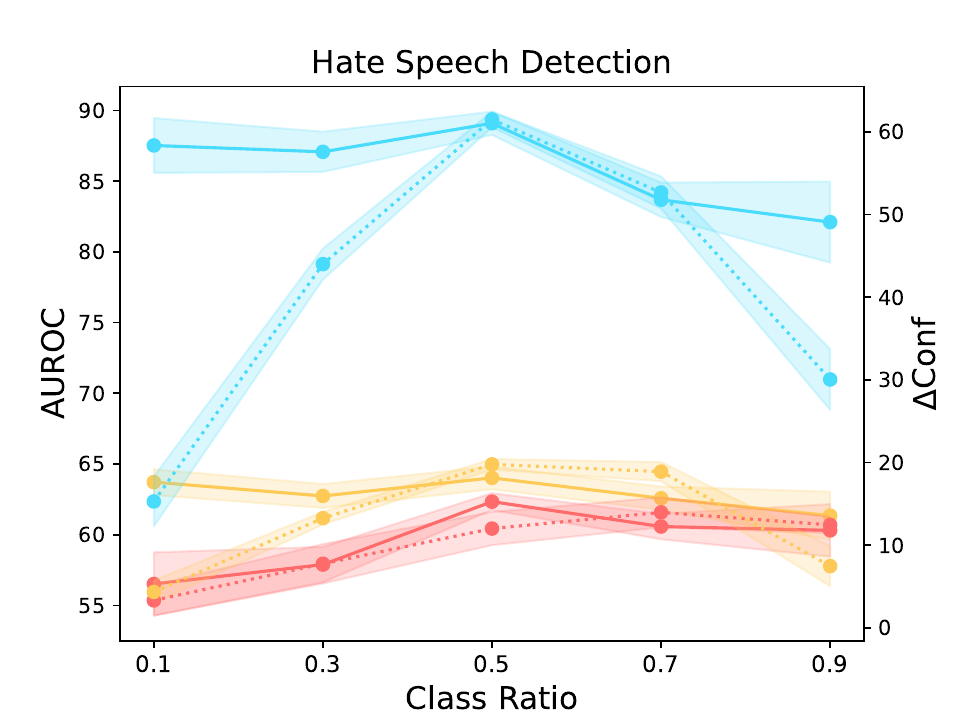}
%   \hspace{1in}
  \includegraphics[width=0.32\textwidth]{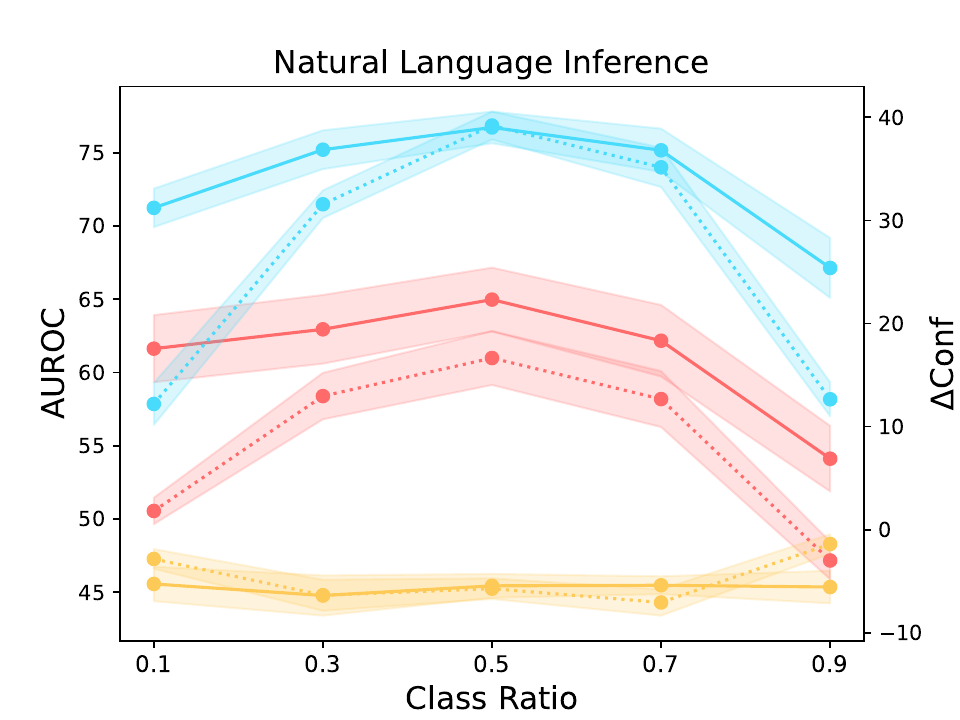}
   \hspace{1in}
   \includegraphics[width=0.95\textwidth]{figs/legend.pdf}
  
  \caption{Quantify the influence of dataset imbalance in the calibration task. The evaluation datasets are listed in Table~\ref{tab:dataset}. An exact balanced distribution of the two classes will mostly benefit the calibration task.}
  \vspace{-15pt}
   \label{fig:quantify_2}
\end{figure*}

%% file: tabs/quantify_3.tex
\begin{table}[]

\centering
\resizebox{0.49\textwidth}{!}{
\begin{tabular}{l|l|rrrrrr}
\toprule
\multicolumn{1}{c|}{\multirow{5}{*}{Amazon}} & Dataset        & \multicolumn{2}{c}{Amazon}                                   & \multicolumn{2}{c}{SST-5}                                    & \multicolumn{2}{c}{SemEval}                                  \\ \cmidrule(l){2-8} 
\multicolumn{1}{c|}{}                        & Method         & \multicolumn{1}{c}{AUROC} & \multicolumn{1}{c}{$\Delta$Conf} & \multicolumn{1}{c}{AUROC} & \multicolumn{1}{c}{$\Delta$Conf} & \multicolumn{1}{c}{AUROC} & \multicolumn{1}{c}{$\Delta$Conf} \\ \cmidrule(l){2-8} 
\multicolumn{1}{c|}{}                        & All Features    & \textbf{89.44 (0.08)}     & \textbf{73.96 (0.35)}            & \textbf{61.95 (0.16)}    & \textbf{28.94 (0.82)}            & \textbf{73.70 (1.18)}     & \textbf{28.65 (0.57)}            \\
\multicolumn{1}{c|}{}                        & w/o Prediction & 84.12 (0.49)              & 64.81 (0.87)                     & 59.47 (1.04)              & 24.46 (0.24)                     & 68.15 (1.42)              & 22.46 (2.67)                     \\
\multicolumn{1}{c|}{}                        & w/o Sample  & 70.06 (1.67)              & 3.93 (1.07)                      & 57.22 (3.31)              & 1.16 (0.55)                      & 59.19 (0.94)              & 1.42 (0.37)                      \\ \midrule
\multirow{5}{*}{Civil}                       & Dataset        & \multicolumn{2}{c}{Civil}                                    & \multicolumn{2}{c}{Hate Speech}                              & \multicolumn{2}{c}{Implicit}                                     \\ \cmidrule(l){2-8} 
                                             & Method         & \multicolumn{1}{c}{AUROC} & \multicolumn{1}{c}{$\Delta$Conf} & \multicolumn{1}{c}{AUROC} & \multicolumn{1}{c}{$\Delta$Conf} & \multicolumn{1}{c}{AUROC} & \multicolumn{1}{c}{$\Delta$Conf} \\ \cmidrule(l){2-8} 
                                             & All Features    & \textbf{86.98 (5.00)}     & \textbf{59.30 (0.10)}           & 60.44 (0.39)             & 11.18 (2.28)                     & \textbf{64.16 (1.98)}     & \textbf{21.05 (2.95)}            \\
                                             & w/o Prediction & 83.89 (2.51)              & 52.74 (3.83)                     & \textbf{61.83 (4.52)}     & \textbf{14.90 (8.28)}            & 63.93 (0.32)              & 20.96 (0.22)                     \\
                                             & w/o Sample  & 5.18 (0.20)               & -3.5 (2.39)                      & 55.73 (0.79)              & 4.35 (0.27)                      & 37.02 (0.21)              & -1.02 (0.71)                     \\ \midrule
\multirow{5}{*}{MNLI}                        & Dataset        & \multicolumn{2}{c}{MNLI}                                     & \multicolumn{2}{c}{HANS}                                     & \multicolumn{2}{c}{ANLI}                                     \\ \cmidrule(l){2-8} 
                                             & Method         & \multicolumn{1}{c}{AUROC} & \multicolumn{1}{c}{$\Delta$Conf} & \multicolumn{1}{c}{AUROC} & \multicolumn{1}{c}{$\Delta$Conf} & \multicolumn{1}{c}{AUROC} & \multicolumn{1}{c}{$\Delta$Conf} \\ \cmidrule(l){2-8} 
                                             & All Features    & 78.80 (0.89)              & 42.64 (0.58)                     & \textbf{67.90 (3.16)}     & \textbf{24.11 (6.55)}            & 44.31 (0.30)     & -6.95 (0.99)          \\
                                             & w/o Prediction & \textbf{79.24 (0.31)}     & \textbf{44.94 (0.42)}            & 60.22 (2.11)              & 14.11 (4.32)                     & 43.99 (0.23)              & -8.05 (1.33)                     \\
                                             & w/o Sample  & 39.19 (0.89)              & 0 (0.11)                         & 39.12 (0.34)              & 0 (0.09)                         & \textbf{51.28 (0.24)}              & \textbf{0 (0.01)}                         \\ \bottomrule
\end{tabular}
}

\caption{Quantify the influence of ``features'' in the calibration task. 
Numbers in parentheses are standard deviations. Both ``features'' contribute to the predictions.}
% \vspace{-31pt}
\label{tab:quantify_3}
\end{table}

%% file: method.tex
\begin{figure*}
  \centering
  \includegraphics[width=0.85\textwidth]{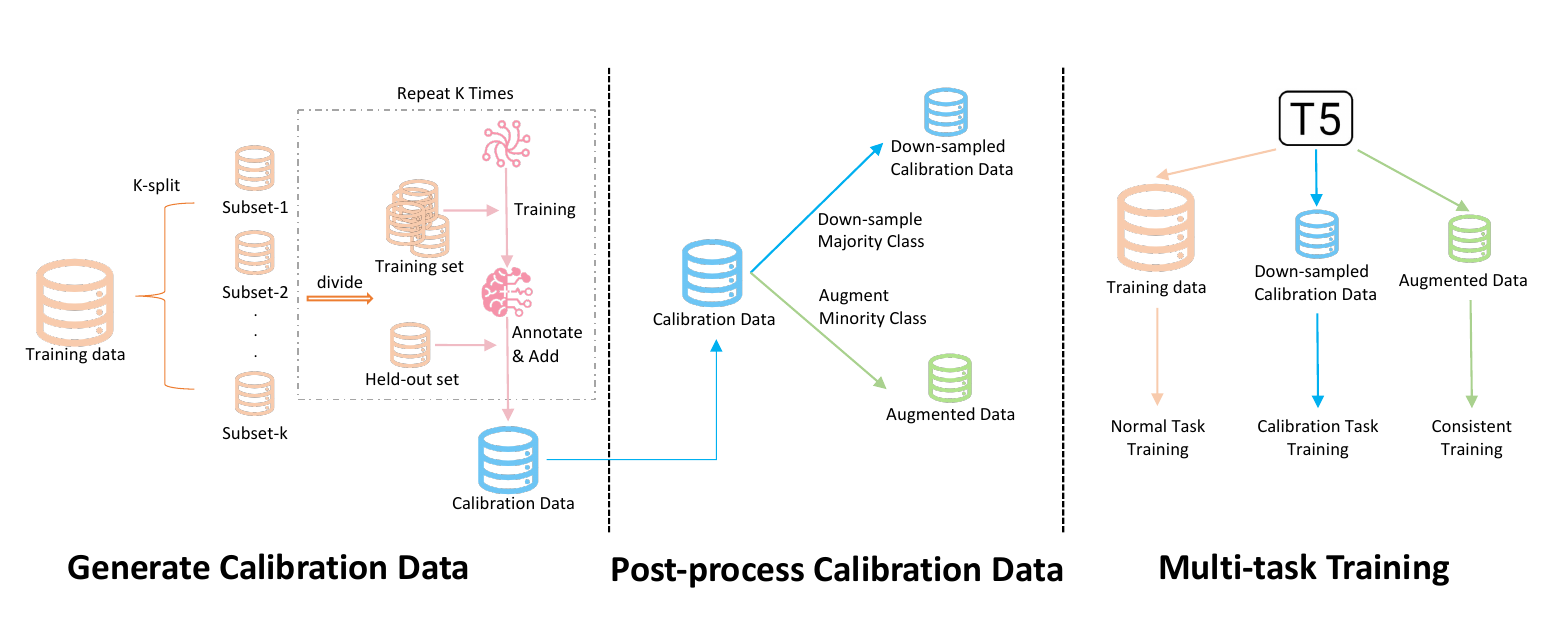}
%   \hspace{1in}
  \vspace{-10pt}
  \caption{The demonstration of LM-TOAST \includegraphics[width=0.017\textwidth]{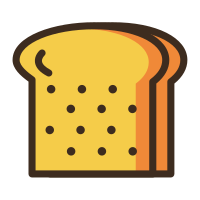}, consisting of three separate stages, namely generating calibration data, post-processing calibration data, and multi-task training.
  }
   \label{fig:method}
\end{figure*}

\section{Method}
Based on our empirical analysis summarized in Appendix~\ref{sec:appendix:summary}, we motivate a practical training algorithm to make PLMs both \textbf{T}ask-s\textbf{O}lvers \textbf{A}nd \textbf{S}elf-calibra\textbf{T}ors (LM-TOAST\includegraphics[width=0.017\textwidth]{figs/toast.png}). 
LM-TOAST can be divided into three stages (see Figure~\ref{fig:method}):
(1) Generate calibration training data from the given training dataset;
(2) Post-process generated calibration training data;
(3) Multi-task training using the original training data and the processed calibration training data. 
\textbf{We follow the notations in Sec.~\ref{sec:task_formalization}. }
% 

% \looseness=-1
\paragraph{Generate calibration training data.}
We propose the K-fold cross-annotation to generate the calibration dataset from the original training samples.
We first split the original training dataset into K subsets equally, and perform K-rounds annotation. 
For each round, we leave one subset out and train the model on the remaining K-1 subsets. 
Then we use the trained model to annotate the held-out set. 
Specifically, for each sample in the held-out set, we obtain the model's prediction and compare it with the golden label. 
A binary annotated label is obtained, indicating whether the prediction is correct or not.
After K-rounds annotation, we can generate a calibration training dataset $\mathbb{D}^*$ with the size equal to the original training dataset's size.
We empirically set K=2 in LM-TOAST to avoid hyper-parameter searching. 
We justify this setting in the further analysis of LM-TOAST (see Appendix~\ref{sec:appendix:further}). 
Note that due to the strong capacity of PLMs, there exists a significant data imbalance issue in $\mathbb{D}^*$, where positive cases dominate the distribution. 
Thus, post-processing the generated calibration data is needed to make full use of $\mathbb{D}^*$.

\paragraph{Post-process generated calibration training data.}
Data imbalance is a long-standing research problem in machine learning. 
We visit existing methods for our calibration task. 
According to our pilot exploration, we adopt two strategies tailored for our task, namely down-sampling the majority class and performing data augmentation on the minority class. 
For the former one, we simply down-sample the majority class to achieve the exact balance of two classes in the calibration training dataset.
For data augmentation on the minority class, we employ textual transformation methods~\citep{DBLP:conf/emnlp/WeiZ19, DBLP:journals/corr/abs-2110-01852} to generate the constructed set $\mathbb{D}_{A}^*$ that contains augmented negative samples:
\begin{equation}
\mathbb{D}_{A}^*=\{(x_i, f(x_i), y_i^*)_{i=1}^{N}\}, \  (x_i, y_i^*, 0) \sim \mathbb{D}^* 
\end{equation}
where $f$ is the textual transformation method. 
Specifically, we consider the following methods and choose them randomly:
(1) Synonym substitution: Exploiting WordNet~\citep{DBLP:journals/cacm/Miller95} to substitute words in original samples with their synonyms; 
(2) Random insertion: Randomly inserting or repeating some words in original samples; 
(3) Random swap: Randomly swapping the order of adjacent words in original samples;
(4) Random deletion: Randomly deleting some words in original samples. 

\paragraph{Multi-task training.}
After the post-processing, we currently possess the original training set $\mathbb{D}$, the generated and then down-sampled calibration training set $\mathbb{D}^*$, the constructed set $\mathbb{D}_{A}^*$ containing augmented negative samples. 
We find that directly mixing $\mathbb{D}^*$ and $\mathbb{D}_{A}^*$ for the calibration task training has minimal or negative effects on the calibration performance. 
The bottleneck substantially lies in the diversity and quality of the augmented samples, which is the central problem in textual data augmentation research. 

\looseness=-1
Thus, we treat $\mathbb{D}^*$ and $\mathbb{D}_{A}^*$ separately and adopt different training strategies. 
In high-quality $\mathbb{D}^*$, we conduct normal classification training:
\begin{equation}
L_c = \textit{CE}(\mathcal{F}(x_i, y_i^*), c_i), \  (x_i, y_i^*, c_i) \sim \mathbb{D}^*, 
\end{equation}
where CE is the cross-entropy loss. 
For easy reference, $x_i$ is the original sample in D, $y_i^*$ is model’s original prediction, and $c_i$
is a binary value, indicating whether the prediction $y_i^*$ is correct or not.
In $\mathbb{D}_{A}^*$ that contains noise, a robust training algorithm is needed to effectively utilize the augmented samples. 
Specifically, we draw inspirations  from~\citet{DBLP:conf/iclr/MiyatoDG17} that consider the problem of textual adversarial training.
They propose to use consistent training to enforce that the predictive probability of the original input is the same as that of the corresponding input with gradient-based perturbations added in the embedding layer.       
Similarly, we propose to constrain the predictive probability of the original samples and corresponding augmented samples: 
\begin{equation}
L_c^* = \textit{KL}(\mathcal{F}(x_i, y_i^*), \mathcal{F}(x_i^*, y_i^*)), \  (x_i, x_i^*, y_i^*) \sim \mathbb{D}^*_{A}, \\ 
\end{equation}
where KL measures the Kullback–Leibler divergence between two distributions. 
Considering the original task, we conduct multi-task training to minimize the loss $L_A$:
\begin{equation}
L_o = \textit{CE}(\mathcal{F}(x_i), y_i), \  (x_i, y_i) \sim \mathbb{D}, 
\end{equation}
\vspace{-15pt}
\label{eq:main_eq}
\begin{equation}
L_A = L_o + L_c + \alpha * L_c^*, \\ 
\end{equation}
where CE is the cross-entropy loss, $L_o$ is the loss of the original task, and $\alpha$ is a hyper-parameter to control the influence of the consistent loss. 
We empirically set $\alpha$ to 0.1 in LM-TOAST to avoid hyper-parameter searching. 
We justify this setting in the further analysis of LM-TOAST (see Appendix~\ref{sec:appendix:further}).

%% file: exp.tex
\section{Experiments}
We conduct experiments to demonstrate the effectiveness of LM-TOAST in confidence estimations. We run all experiments three times and report both the average performance and the standard variance. 

\label{sec:exp}
\subsection{Baseline Methods}
\looseness=-1
We adopt three baseline methods for confidence estimations:
(1) \textbf{Vanilla}: Use the original predictive probability as the confidence estimation; 
(2) \textbf{Temperature Scaling (TS)}: Apply the temperature scaling method to calibrate PLMs' confidence scores~\citep{DBLP:conf/icml/GuoPSW17}; 
(3) \textbf{Label Smoothing (LS)}: Apply label smoothing to prevent PLMs from becoming overconfident in their predictions~\citep{DBLP:conf/cvpr/SzegedyVISW16}.

\subsection{Results of Calibration Performance}
\input{tabs/main_exp}

\looseness=-1
The experimental results are listed in Table~\ref{tab:main_exp}. 
We observe that LM-TOAST achieves overall better calibration performance. 
For fine-grained confidence ranking of correct and wrong predictions (AUROC), LM-TOAST improves the discrimination of wrong predictions by assigning them with relatively lower confidence. 
Also, we note that vanilla confidence can already be adopted for effectively detecting wrong predictions, consistent with previous work~\citep{DBLP:conf/acl/HendrycksLWDKS20}.

However, as shown in Figure~\ref{fig:hello}, there is no distinguished confidence gap between confidence scores on correct and wrong predictions.
This results in many high-confident wrong predictions, which is undesirable since false acceptance may happen in reality. 
Besides, previous work regarding the utilization of vanilla confidence scores (e.g., selective classification~\citep{DBLP:conf/acl/KamathJL20}) overlooks this issue since the selected metric mostly doesn't need a chosen threshold (e.g., AUROC).
But in practice, a small confidence gap between correct and wrong predictions makes it hard for practitioners to manually select a concrete threshold and may cause large performance variance. 
Thus, it's also essential to measure the confidence gap between correct and wrong predictions. 
We show that LM-TOAST can significantly increase this gap, in both ID and OOD settings. 
One exception is still the ANLI dataset. 
We refer to Sec.~\ref{sec:num_train} for our explanation. 

Due to space limits, we present further analysis of LM-TOAST in Appendix~\ref{sec:appendix:further}, including the ablation study of each component and the influence of the hyper-parameter K in the cross-annotation.

%% file: tabs/main_exp.tex
\begin{table}[]

\centering
\resizebox{0.5\textwidth}{!}{
\begin{tabular}{l|l|rrrrrr}
\toprule
\multicolumn{1}{c|}{\multirow{6}{*}{Amazon}} & Dataset  & \multicolumn{2}{c}{Amazon}                                   & \multicolumn{2}{c}{SST-5}                                    & \multicolumn{2}{c}{SemEval}                                  \\ \cmidrule(l){2-8} 
\multicolumn{1}{c|}{}                        & Method   & \multicolumn{1}{c}{AUROC} & \multicolumn{1}{c}{$\Delta$Conf} & \multicolumn{1}{c}{AUROC} & \multicolumn{1}{c}{$\Delta$Conf} & \multicolumn{1}{c}{AUROC} & \multicolumn{1}{c}{$\Delta$Conf} \\ \cmidrule(l){2-8} 
\multicolumn{1}{c|}{}                        & Vanilla  & 85.80 (0.45)              & 15.18 (0.33)                     & 79.14 (0.83)              & 16.56 (0.78)                     & 71.68 (0.93)              & 12.68 (0.59)                     \\
\multicolumn{1}{c|}{}                        & TS       & 85.80 (0.45)              & 16.85 (0.59)                     & 79.14 (0.83)              & 10.63 (0.34)                     & 71.68 (0.93)              & 8.07 (0.64)                      \\
\multicolumn{1}{c|}{}                        & LS       & 81.93 (2.77)              & 13.81 (0.83)                     & 76.81 (1.41)              & 13.45 (0.99)                     & 70.52 (1.26)              & 10.81 (0.90)                     \\
\multicolumn{1}{c|}{}                        & LM-TOAST \includegraphics[width=0.017\textwidth]{figs/toast.png} & \textbf{87.44 (1.12)}     & \textbf{42.46 (3.03)}            & \textbf{79.32 (1.09)}     & \textbf{26.37 (1.04)}            & \textbf{73.17 (2.36)}     & \textbf{20.21 (1.70)}            \\ \midrule
\multirow{6}{*}{Civil}                       & Dataset  & \multicolumn{2}{c}{Civil}                                    & \multicolumn{2}{c}{Hate Speech}                              & \multicolumn{2}{c}{Implicit}                                     \\ \cmidrule(l){2-8} 
                                             & Method   & \multicolumn{1}{c}{AUROC} & \multicolumn{1}{c}{$\Delta$Conf} & \multicolumn{1}{c}{AUROC} & \multicolumn{1}{c}{$\Delta$Conf} & \multicolumn{1}{c}{AUROC} & \multicolumn{1}{c}{$\Delta$Conf} \\ \cmidrule(l){2-8} 
                                             & Vanilla  & 90.33 (0.99)              & 14.23 (0.09)                     & 62.80 (0.55)              & 2.74 (0.64)                      & \textbf{65.99 (1.22)}     & 3.63 (1.19)                      \\
                                             & TS       & 90.33 (0.99)              & 22.71 (0.46)                     & 62.80 (0.55)              & 5.60 (0.27)                      & \textbf{65.99 (1.22)}     & 7.25 (0.23)                      \\
                                             & LS       & 91.15 (0.11)              & 11.59 (0.32)                     & 62.17 (0.22)              & 2.08 (0.33)                      & 63.92 (0.13)              & 3.30 (0.34)                      \\
                                             & LM-TOAST \includegraphics[width=0.017\textwidth]{figs/toast.png} & \textbf{92.01 (0.14)}     & \textbf{51.38 (0.24)}            & \textbf{65.55 (1.76)}     & \textbf{12.57 (2.47)}            & \textbf{65.99 (0.12)}     & \textbf{17.29 (0.30)}            \\ \midrule
\multirow{6}{*}{MNLI}                        & Dataset  & \multicolumn{2}{c}{MNLI}                                     & \multicolumn{2}{c}{HANS}                                     & \multicolumn{2}{c}{ANLI}                                     \\ \cmidrule(l){2-8} 
                                             & Method   & \multicolumn{1}{c}{AUROC} & \multicolumn{1}{c}{$\Delta$Conf} & \multicolumn{1}{c}{AUROC} & \multicolumn{1}{c}{$\Delta$Conf} & \multicolumn{1}{c}{AUROC} & \multicolumn{1}{c}{$\Delta$Conf} \\ \cmidrule(l){2-8} 
                                             & Vanilla  & 82.08 (0.43)              & 12.25 (0.19)                     & 51.14 (1.68)              & 2.59 (0.52)                      & \textbf{44.14 (0.31)}     & \textbf{-2.31 (0.06)}            \\
                                             & TS       & 82.08 (0.43)              & 18.30 (0.32)                     & 51.14 (1.68)              & -1.87 (0.70)                     & \textbf{44.14 (0.31)}     & -2.89 (0.13)                     \\
                                             & LS       & 80.51 (0.14)              & 10.12 (0.39)                     & 46.52 (2.32)              & 0.53 (0.80)                      & 43.07 (0.32)              & -2.91 (0.92)                     \\
                                             & LM-TOAST \includegraphics[width=0.017\textwidth]{figs/toast.png} & \textbf{82.74 (0.49)}     & \textbf{33.53 (1.40)}            & \textbf{60.60 (4.34)}     & \textbf{11.19 (3.81)}            & 43.97 (0.62)              & -6.69 (0.70)                     \\ \bottomrule
\end{tabular}
}
% \vspace{-pt}

\caption{Experimental results of calibration performance. Numbers in parentheses are standard deviations.}
\label{tab:main_exp}
\end{table}

%% file: application.tex
\section{Applications}
We consider applying LM-TOAST for three tasks, namely selective classification~\citep{geifman2017selective}, adversarial defense~\citep{DBLP:journals/tist/ZhangSAL20}, and model cascading~\citep{varshney2022investigating}.
The baseline methods are the same as in Sec.~\ref{sec:exp}

\subsection{Selective Classification}

Selective classification provides systems with an extra reject option.
It plays an essential role in high-stake application scenarios (e.g., fake news detection) since the systems can trade off the prediction coverage for better prediction performance.
Once PLMs' confidence scores on predictions are lower than the pre-defined threshold, the systems may reject PLMs' predictions and transfer the inputs to human experts. 
Thus, the task performance can be improved by clearly distinguishing the wrong predictions. 
The evaluation settings are described in Appendix~\ref{sec:appendix:eval_metric}.

% \looseness=-1
% \paragraph{Evaluation metrics.} 

% \textbf{Metric1: AUC}. Already have, inverse. 
% \textbf{Metric2: the maximum possible coverage for a desired risk level. e.g., when acc=95\%}

\paragraph{Experimental results.}
The results are listed in Table~\ref{tab:app1}.
We observe that LM-TOAST overall achieves both the minimum risk averaging over various coverage measured by AUROC$_{risk}$ and the maximum coverage for the desired risk level. 
To further show the advantage of LM-TOAST, we plot the accuracy versus confidence level (a.k.a., threshold) curves for three ID datasets (see Figure~\ref{fig:app1}). 
The predictions with confidence scores lower than the confidence level will be rejected. 
We observe that LM-TOAST can steadily increase performance when the confidence level keeps getting larger.
Besides, LM-TOAST achieves an overall better balance between accuracy and coverage rate. 
On the contrary, while using TS can achieve good performance in high confidence levels, the coverage rate is very low (near 0 on Amazon).

\input{figs/app2}

\subsection{Adversarial Defense}
\label{sec:adv_defense}
\input{tabs/app2}

\looseness=-1
For PLMs deployed for security-relevant applications (e.g., hate-speech detection), malicious agents may construct adversarial samples to mislead PLMs' predictions.
Essentially, the attack methods introduce noise to the original samples, which may result in various degrees of distribution shifts~\citep{DBLP:conf/nips/ZhangZL15, DBLP:journals/corr/abs-2110-15317, DBLP:conf/naacl/0002WY22}. 
Thus, adversarial sample detection can be treated as a special kind of OOD detection problem, and the confidence scores can be exploited. 
The intuition is that PLMs' confidence scores may be lower in adversarial OOD samples compared to ID ones. 
The evaluation settings are described in Appendix~\ref{sec:appendix:eval_metric}. Basically, a sample is considered adversarial when the predictive probability is below a certain threshold.

% \paragraph{Evaluation metrics.} 

\paragraph{Experimental results.}
The results are listed in Table~\ref{tab:app2}. 
We observe that LM-TOAST achieves significantly better performance in detecting adversarial inputs. 
Further, we sample 1,000 sentences from the ID dataset and mix them with adversarial samples.
We measure the Macro-F1 score at various confidence levels considering five attack methods (see Figure~\ref{fig:adversarial_defense}). 
The samples with confidence scores lower than the confidence level will be treated as adversarial samples. 
We observe that LM-TOAST reacts more actively to the confidence threshold chosen, and consistently achieves better detection results across all thresholds. 

\input{tabs/app3}

\input{figs/app3}

\subsection{Model Cascading}
Model cascading systems build a model pool consisting of PLMs with various scales~\citep{varshney2022model}. 
Given input samples in the inference time, smaller models can be first adopted for predictions. 
If the predictive confidence scores are relatively lower, the system can transfer samples to larger models, which will take more time to solve but with more accuracy. 
The basic intuition is that smaller models can already give correct predictions in most cases, and larger models are only needed to be adopted when solving some difficult samples. 
In this way, model cascading systems can significantly improve the efficiency in the inference time. 
The evaluation settings are described in Appendix~\ref{sec:appendix:eval_metric}.

% The current bottleneck in this cascading framework lies in the unreasonable confidence estimations in PLMs. 
% % 
% Wrong predictions are often given high confidence scores, resulting in poor performance of the framework.
% % 
% Thus, our method can contribute to a more reasonable confidence estimations, and make the transfer decision truly reflect the actual cases without hurting the performance.

% \looseness=-1
\paragraph{Experimental results.}
The results are listed in Table~\ref{tab:app3}. 
LM-TOAST achieves better performance on three datasets.
Thus, incorporating the confidence estimations computed by LM-TOAST can improve the efficiency and performance of the cascading systems. 
% 
% Besides, we note that in more challenging real-world inputs that smaller models cannot handle well, LM-TOAST can bring much more benefits.  
% 
We also show the accuracy versus the confidence level curves in Figure~\ref{fig:app3}. 
The samples with confidence scores from the small model lower than the confidence level will be transferred to the large model for prediction.
LM-TOAST still exhibits the benefit of reacting dynamically with the confidence changing and consistently achieves better performance considering all thresholds.

%% file: figs/app2.tex
\begin{figure*}
  \centering
  \includegraphics[width=0.19\textwidth]{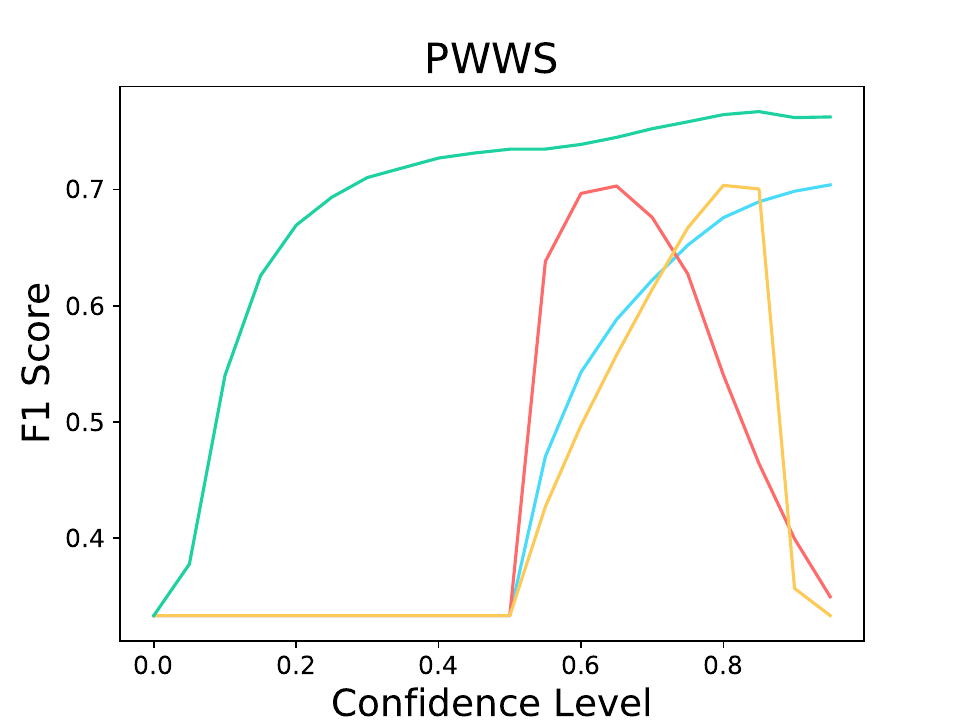}
%   \hspace{1in}
  \includegraphics[width=0.19\textwidth]{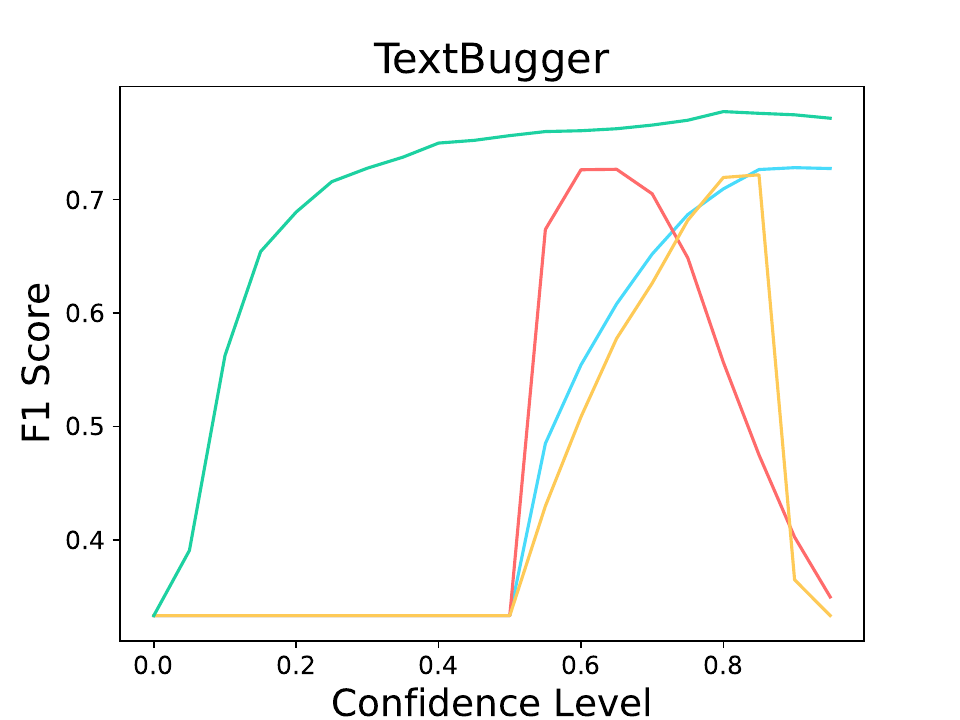}
%   \hspace{1in}
  \includegraphics[width=0.19\textwidth]{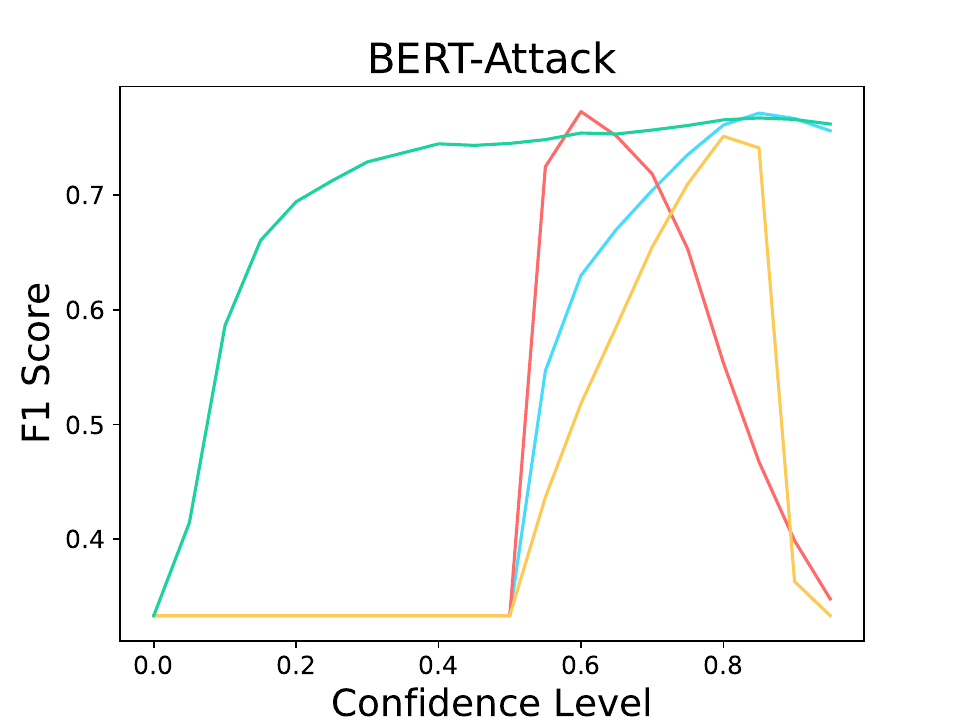}
  \includegraphics[width=0.19\textwidth]{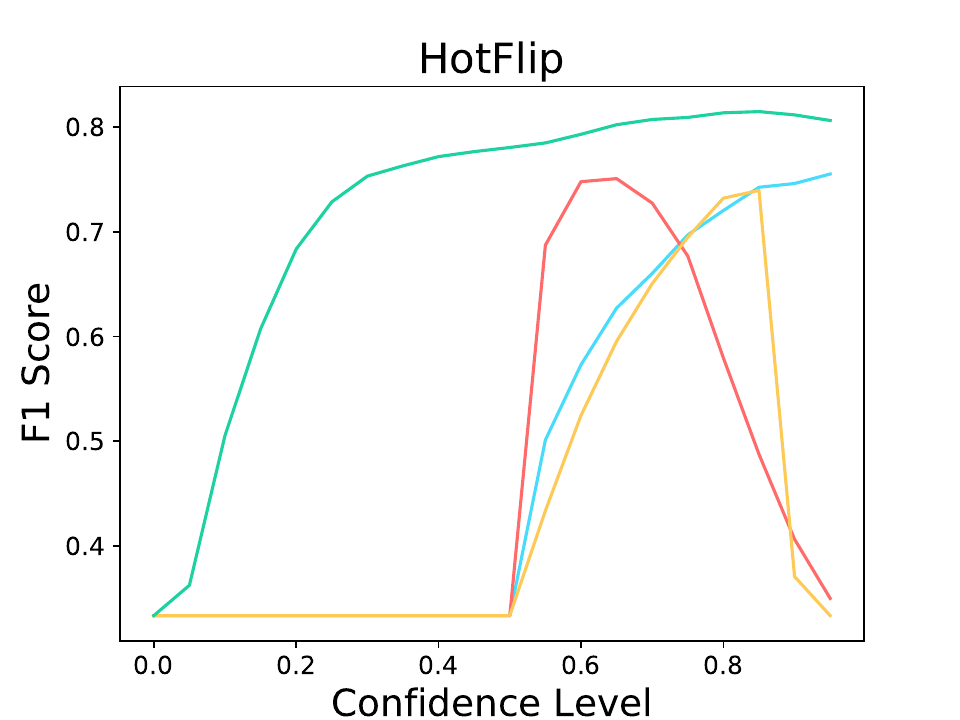}
  \includegraphics[width=0.19\textwidth]{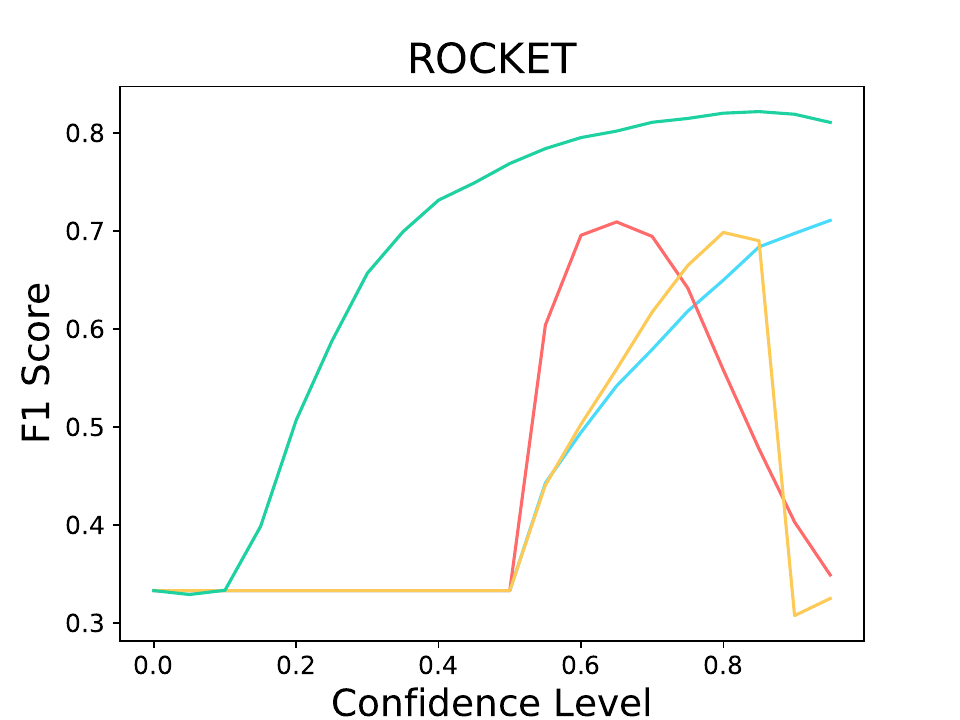}
  \hspace{1in}
  \includegraphics[width=0.7\textwidth]{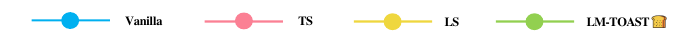}
  \vspace{-10pt}
  \caption{Results of adversarial defense against five classic attack methods. LM-TOAST consistently achieves better detection results across all confidence thresholds.}
   \label{fig:adversarial_defense}
\end{figure*}

%% file: tabs/app2.tex
\begin{table}[]

\centering
\resizebox{0.5\textwidth}{!}{
\begin{tabular}{l|cccccccccc}
\toprule
Dataset  & \multicolumn{2}{c}{PWWS}        & \multicolumn{2}{c}{TextBugger}  & \multicolumn{2}{c}{BERT-Attack} & \multicolumn{2}{c}{HotFlip}     & \multicolumn{2}{c}{ROCKET}      \\ \midrule
Method   & AUROC          & $\Delta$Conf   & AUROC          & $\Delta$Conf   & AUROC          & $\Delta$Conf   & AUROC          & $\Delta$Conf   & AUROC          & $\Delta$Conf   \\ \midrule
Vanilla  & 74.91          & 15.66          & 79.11          & 17.68          & 81.13          & 21.57          & 83.54          & 19.01          & 77.23          & 14.24          \\
TS       & 74.91          & 10.79          & 79.11          & 12.53          & 81.13          & 13.79          & 83.54          & 14.19          & 77.23          & 11.33          \\
LS       & 71.10          & 10.28          & 74.53          & 11.18          & 76.51          & 12.31          & 78.69          & 12.15          & 64.24          & 9.26           \\
LM-TOAST \includegraphics[width=0.017\textwidth]{figs/toast.png} & \textbf{84.64} & \textbf{42.66} & \textbf{86.01} & \textbf{45.75} & \textbf{84.59} & \textbf{45.08} & \textbf{87.96} & \textbf{49.56} & \textbf{85.83} & \textbf{42.78} \\ \bottomrule
\end{tabular}
}
\caption{Experimental results of adversarial defense.}
% \vspace{-20pt}
\label{tab:app2}
\end{table}

%% file: tabs/app3.tex
\begin{table}[]

\centering
\resizebox{0.4\textwidth}{!}{
\begin{tabular}{l|ccc}
\toprule
Dataset  & Amazon         & Civil          & MNLI           \\ \midrule
Method   & AUROC          & AUROC          & AUROC          \\ \midrule
Vanilla  & 88.24          & 86.55          & 81.96          \\
TS       & 88.37          & 86.60          & 82.21          \\
LS       & 88.34          & 86.62          & 82.30          \\
LM-TOAST \includegraphics[width=0.017\textwidth]{figs/toast.png} & \textbf{89.50} & \textbf{88.54} & \textbf{83.93} \\ \bottomrule
\end{tabular}
}
\caption{Experimental results of model cascading.}
% \vspace{-20pt}
\label{tab:app3}
\end{table}

%% file: figs/app3.tex
\begin{figure*}
  \centering
  \includegraphics[width=0.27\textwidth]{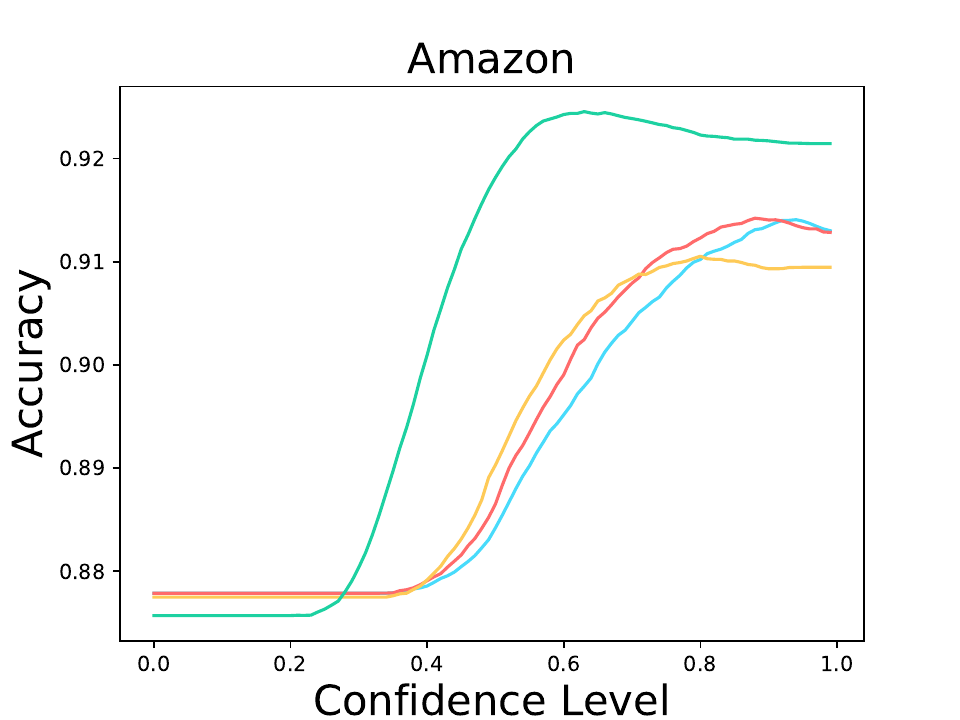}
%   \hspace{1in}
  \includegraphics[width=0.27\textwidth]{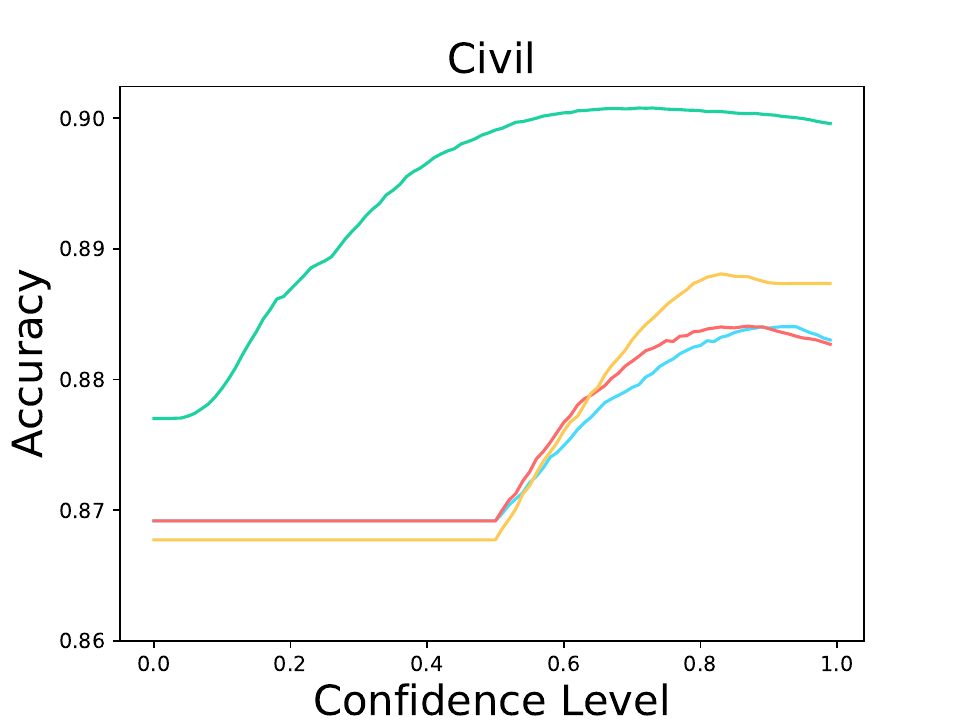}
%   \hspace{1in}
  \includegraphics[width=0.27\textwidth]{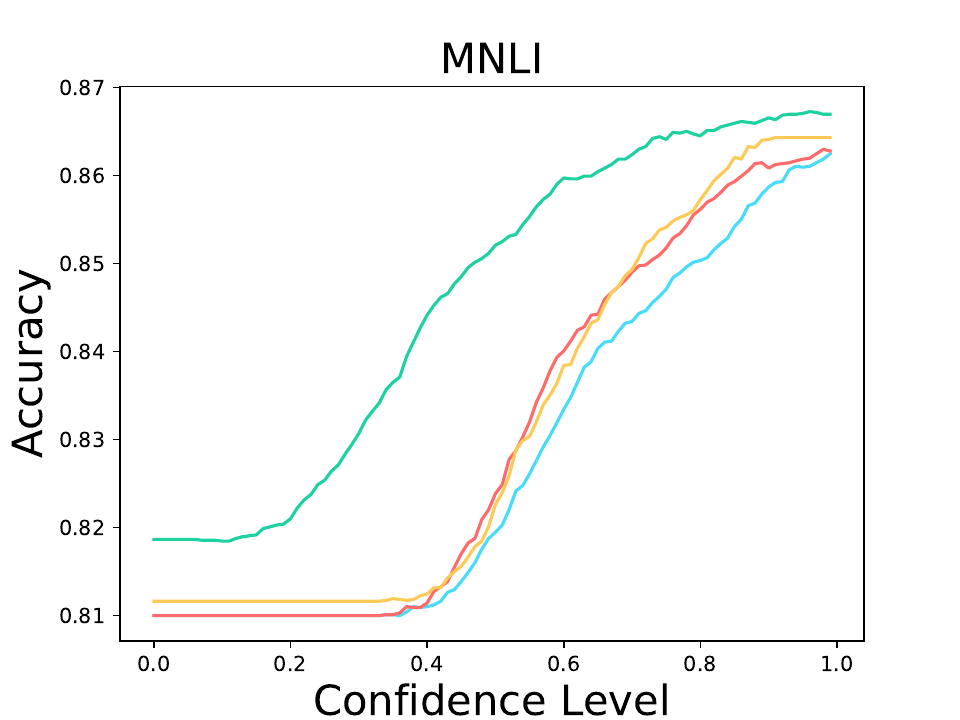}
   \hspace{1in}
   \includegraphics[width=0.7\textwidth]{figs/legend3.pdf}
  \vspace{-10pt}
  \caption{Results of model cascading. LM-TOAST performs consistently better for all confidence thresholds.}
  \vspace{-5pt}
   \label{fig:app3}
\end{figure*}

%% file: related.tex
\section{Related Work}
% \vspace{-10pt}
\paragraph{Calibration methods.}
\looseness=-1
Typically, calibration methods rely on human intuitions or posterior adjustments to make the confidence estimations more accurate.
Data augmentation~\citep{DBLP:conf/iclr/HendrycksMCZGL20, DBLP:conf/nips/WangXKYAW21} and model ensemble~\citep{DBLP:conf/icml/GalG16, DBLP:conf/nips/Lakshminarayanan17} have been empirically proven to be successful in computer vision.
However, they cannot bring the same benefits in NLP according to the empirical study in~\citet{DBLP:journals/corr/abs-2211-00151}.
So we don't consider them as baseline methods for comparison in this work, and two empirically effective methods are adopted. 
Temperature scaling~\citep{platt1999probabilistic, DBLP:conf/icml/GuoPSW17} readjusts the output logits in a posterior way according to the calibration performance on a held-out set. 
Label smoothing~\citep{DBLP:conf/cvpr/SzegedyVISW16} imposes a confidence regularization during the training process, discouraging models from being overconfident in their predictions. 

Recently, there is an emergent trend in NLP, motivating to directly collect data for training models to have reasonable confidence estimations.
% 
% Most existing research is based on PLMs. 
% 
\citet{kadavath2022language} assume that the last hidden states of PLMs contain the uncertainty information, and directly apply a multi-layer perceptron on them to perform confidence estimations. 
\citet{lin2022teaching} also show that PLMs can be directly trained to give their confidence estimations by words.
These two methods are proven to be successful in significantly reducing the overconfidence issue and are further extended to exploit the potential in this kind of methods~\citep{DBLP:journals/corr/abs-2211-00151}. 
Existing work demonstrates the feasibility and potential of this kind of method in ideal experimental settings that enough training data is given for the calibration task.
We further consider the practical setting and propose an effective method in this work.
% and propose a method that can effectively utilize a limited number of training data for both the main task and the calibration task. 

% Then train a model to predict whether the given answer is correct. 
% The model can be a multilayer perceptron, and the features can be hand-engineered~\citep{DBLP:conf/acl/YeD22, DBLP:conf/acl/ZhangGC21, si2022revisiting} or the last hidden states of PLMs

% \vspace{-30pt}

\paragraph{Applications.}
The confidence scores have been widely utilized for various applications. 
A bunch of active learning methods relies on models' confidence to select the most informative samples to annotate~\citep{DBLP:journals/corr/abs-2210-10109, DBLP:conf/acl/SchroderNP22}. 
Models' confidence can also be directly utilized for OOD and misclassification detection~\citep{DBLP:conf/iclr/HendrycksG17, DBLP:conf/acl/HendrycksLWDKS20}.
Following the same intuition, selective prediction can be applied to improve the system performance by filtering out low-confident predictions~\citep{geifman2017selective, DBLP:conf/acl/KamathJL20, varshney2022investigating}.
Moreover, an alternative strategy is adopting the model cascading systems, transferring the low-confident inputs to models with higher capacities~\citep{DBLP:conf/emnlp/LiLCRLZS21, varshney2022model}. 
This can achieve better performance and efficiency of the whole system.

% selective prediction~\citep{varshney2022investigating}, robustness~\citep{kumar2022calibrated}, and pseudo-labeling~\citep{rizve2021defense}. 

%% file: limitation.tex
\section*{Limitations and Future Work}
We acknowledge the limitations of LM-TOAST in few-shot calibration training.
From our pilot experiments in Sec.~\ref{sec:num_train}, we observe that a significant amount of data points are needed for the calibration task training.
In LM-TOAST, we effectively utilize the whole training set for the calibration task.
Some learning paradigms assume only a small number of annotated samples at first, which limits the effectiveness of LM-TOAST.
For example, in active learning~\citep{DBLP:journals/corr/abs-2210-10109, DBLP:conf/acl/SchroderNP22}, only a very small number of samples are available at the beginning most of the time, and models need to rely on them to find informative unlabeled samples to annotate. 
The vanilla confidence scores can be effectively utilized in this setting. 
However, LM-TOAST may not learn the calibration task well given very limited samples, resulting in poor performance in searching informative samples. 
We plan to investigate the calibration task in the few-shot setting and bring out the potential in LM-TOAST to make it suitable for more tasks.

%% file: appendix.tex
% \quad

% \newpage
\input{tabs/dataset_statistics}

% \section*{Appendix}
\section{Dataset}
\label{appendix:dataset}

We introduce the datasets used in this paper. 
The dataset statistics are listed in Table~\ref{tab:dataset_stat}.
Most datasets are selected from the BOSS benchmark~\cite{DBLP:journals/corr/abs-2306-04618}.

\paragraph{Sentiment analysis.}
We choose \textbf{Amazon Fine Foods}~\citep{amazon2013mcauley}, abbreviated as Amazon in this paper, as the ID dataset. 
It collects customer reviews from Amazon on fine foods. 
Following~\citet{DBLP:journals/corr/abs-2211-00151}, we sample 10k samples per class due to the enormous size of the original dataset.
For OOD datasets, we choose \textbf{SST-5}~\citep{sst2013socher} and \textbf{SemEval} 2016 Task 4 \citep{semeval2016nakov} for evaluation. 
Specifically, SST-5 collects samples from the movie reviews website, and all samples are annotated using 5 sentiment tendencies, including negative, somewhat negative, neutral, somewhat positive, or positive.
We discard the samples with somewhat positive and somewhat negative labels and make it a three-classes classification dataset. 
SemEval collects samples from Twitter, where each sample is annotated as negative, neutral, or positive.

\paragraph{Hate speech detection.}
We choose \textbf{Civil Comments}\footnote{\url{https://www.kaggle.com/competitions/jigsaw-unintended-bias-in-toxicity-\\classification}}, abbreviated as Civil in this paper, as the ID dataset. 
It collects samples from the Civil Comments platform, and each sample is annotated as a value from 0 to 1, indicating the toxicity level. 
Following the official instructions, we set the samples with toxicity levels larger than 0.5 as toxic labels and smaller than 0.5 as benign labels.
For OOD datasets, we choose \textbf{Hate Speech} \citep{hatespeech2018de-gibert} and \textbf{Implicit Hate}~\citep{implicithate2021elsherief}, abbreviated as Implicit in this paper, for evaluation. 
Hate Speech collects samples from a white nationalist forum. 
We use the test set sampled in the official repository. 
Implicit collects tweets that contain toxic content from extremist groups in the US. 
This dataset is challenging for hate speech detection since many samples contain implicit toxic contents to bypass the detection systems.

\paragraph{Natural language inference.}
We choose \textbf{MNLI}~\citep{mnli2018williams} as the ID dataset. 
In our experiments, we choose the matched version validation set for evaluation. 
For OOD datasets, we choose \textbf{HANS} \citep{hans2019mccoy} and \textbf{ANLI} \citep{anli2020nie} for evaluation. 
HANS is a synthetic dataset, constructed by heuristic rules. 
It's used to evaluate whether models trained on standard natural language inference datasets capture some unwanted spurious correlations. 
ANLI is construed by a human-and-model-in-the-loop process, aiming to explore the weakness of standard natural language inference models. 
In our experiments, we merge the data from three rounds in the original ANLI dataset. 

\section{Prompt Template and Verbalizer}
\label{sec:appendix:template}
% \looseness=-
We follow~\citet{DBLP:journals/corr/abs-2211-00151} to select the prompt templates and verbalizers chosen. 
We list the prompt templates and verbalizers for the main task in Table~\ref{tab:template}.
We list the prompt templates and verbalizers for the calibration task in Table~\ref{tab:template2}.

\section{Additional Details of the Pilot Experiments}
\label{sec:appendix:summary}
\looseness=-1
\paragraph{Experimental settings.}
We run all experiments three times and report both the average performance and the standard variance. 
Following~\citet{DBLP:journals/corr/abs-2211-00151}, we employ a large unused validation set for pilot exploration, which may not exist in practice. 
We train PLMs on the main task for 5 epochs and use the trained PLMs to annotate the validation set as the calibration training dataset. 
Then we train PLMs on both the main task and the calibration task for 8 epochs.

\paragraph{Summary.}
We summarize the empirical results in the pilot experiments section as instructions for our proposed algorithm:
(1) Increasing the calibration training samples brings benefits to both ID and OOD evaluation;
(2) It's essential to maintain the balanced distribution of two classes in the calibration training set;
(3) Both the original sample and the model's original prediction can provide useful information for confidence estimation, while the former accounts for the majority.

\input{figs/quantify_further}

\section{Further Analysis of LM-TOAST}
\label{sec:appendix:further}
\paragraph{Ablation study.}
\input{tabs/ablation}

We quantify the contribution of each part in LM-TOAST. 
Specifically, we consider four variants of LM-TOAST:
(1) w/o Cross-annotation: We directly split the training dataset into two subsets with a ratio of 9:1, and use the smaller subset for calibration training data annotation; 
(2) w/o Down-sampling: We retain all positive cases in the calibration training set. 
(3) w/o Augmentation: We remove the data augmentation on negative cases in LM-TOAST; 
(4) w/o Decay $\alpha$: We remove the decay factor $\alpha$ in Eq.~\ref{eq:main_eq}.

The results are listed in Table~\ref{tab:ablation}. 
We find that each component in LM-TOAST contributes to a certain aspect of the calibration performance.
The down-sampling, augmentation, and decay factor $\alpha$ guarantee the ID calibration performance, where removing any of them will cause a significant drop in performance. 
Cross-annotation ensures that the amount of calibration training data is large enough, which is important for the OOD calibration performance. 
We also observe that removing the decay factor $\alpha$ improves the OOD calibration performance. 
However, it comes at a substantial cost to the ID performance.

\paragraph{The influence of K.}
We also study the influence of the hyper-parameter K in the cross-annotation process. 
The results are shown in Figure~\ref{fig:ablation}. 
% \
We observe that increasing K brings a negative or minimal effect on the calibration performance. 
Also, increasing K will also cause a significant increase in computational cost. 
Thus, the results justify our empirically chosen value K=2.

\input{figs/app1.tex}

\section{Additional Results}
The results of further analysis of the data imbalance issue are shown in Figure~\ref{fig:quantify_further}.
The results of selective classification are listed in Table~\ref{tab:app1} and Figure~\ref{fig:app1}.

\section{Evaluation Settings of Downstream Applications}
\label{sec:appendix:eval_metric}
\paragraph{Selective classification.}
We consider two classic metrics following~\citet{DBLP:conf/acl/KamathJL20}:
(1) AUROC$_{risk}$: We plot the risk versus coverage curve by varying the confidence threshold, and measuring the area under this curve.
In fact, this can be computed by subtracting the AUROC scores listed in Table~\ref{tab:main_exp} from 1. 
% 
% \xw{maybe a different name to avoid confusion?}
\textbf{Notably, smaller AUROC$_{risk}$ scores are better in selective classification, which is different from other applications}; 
(2) Cov: The maximum possible coverage for the desired risk level. 
We choose Acc=95\% for most tasks while choosing 85\% and 60\% for Semeval and HANS respectively due to PLMs' low performance on these two datasets.
We leave out the results in ANLI due to the same low-performance reason.

\input{figs/ablation}

\paragraph{Adversarial defense.}

In our experiments, we follow~\citet{DBLP:journals/corr/abs-2210-10683} to consider the security-relevant task in evaluation, and use civil comments as the evaluation dataset.
We consider five attack methods, including PWWS~\citep{DBLP:conf/acl/RenDHC19}, Textbugger~\citep{DBLP:conf/ndss/LiJDLW19}, BERT-Attack~\citep{DBLP:journals/corr/abs-2004-09984}, Hotflip~\citep{DBLP:conf/acl/EbrahimiRLD18}, and ROCKET~\citep{DBLP:journals/corr/abs-2210-10683}. 
For each method, we generate 1,000 successful adversarial samples by attacking a well-trained T5 model.

We adopt two evaluation metrics:
(1) AUROC: We measure whether the confidence in ID samples is higher than in adversarial samples;
(2) $\Delta$Conf: The average confidence difference between ID samples and adversarial samples.

It is worth highlighting that our approach can readily be extended to detect poisoned samples~\citep{DBLP:conf/naacl/WallaceZFS21, DBLP:journals/corr/abs-2305-02394}, as they are generated by introducing synthetic noise to benign samples, akin to adversarial samples~\citep{DBLP:conf/nips/CuiYHCLS22}.

\paragraph{Model cascading.}
In our experiments, we use a T5-small and a T5-base to constitute the model pool. 
Following~\citet{DBLP:journals/corr/abs-2210-05528}, we measure the AUROC score in our experiments.
Specifically, we vary the confidence levels, corresponding to different computational costs of the system. 
Then we plot the accuracy versus confidence curve and measure the area under this curve as the AUROC score.

\input{tabs/app1}

\input{tabs/prompt}
\input{tabs/template2}

%% file: tabs/dataset_statistics.tex
\begin{table*}[htp!]
\centering
\resizebox{0.8\linewidth}{!}{
\begin{tabular}{llccccc}
\toprule
Task                                        & Dataset     & \multicolumn{1}{l}{\#Class} & \multicolumn{1}{l}{Average Length} & \multicolumn{1}{l}{\#Train} & \multicolumn{1}{l}{\#Dev} & \multicolumn{1}{l}{\#Test} \\ \midrule
\multirow{3}{*}{Sentiment Analysis}         & Amazon      & 3                           & 77.86                              & 24000                       & 78741                     & 91606                      \\
                                            & SST-5       & 3                           & 18.75                              & -                           & -                         & 1067                       \\
                                            & SemEval     & 3                           & 19.61                              & -                           & -                         & 6000                       \\ \midrule
\multirow{3}{*}{Hate Speech Detection}      & Civil       & 2                           & 52.86                              & 48000                       & 12000                     & 60000                      \\
                                            & Hate Speech & 2                           & 21.55                              & -                           & -                         & 478                        \\
                                            & Implicit    & 2                           & 17.34                              & -                           & -                         & 21479                      \\ \midrule
\multirow{3}{*}{Natural Language Inference} & MNLI        & 3                           & 19.36/10.06                        & 373067                      & 19635                     & 9815                       \\
                                            & HANS        & 2                           & 9.15/5.61                          & -                           & -                         & 30000                      \\
                                            & ANLI        & 3                           & 54.40/10.34                        & -                           & -                         & 3200                       \\ \bottomrule
\end{tabular}
}
\vspace{-5pt}
\caption{Dataset statistics.}
% \vspace{-15pt}
\label{tab:dataset_stat}
\end{table*}

%% file: figs/quantify_further.tex
% \vspace{-10pt}
\begin{figure*}
  \centering
  \includegraphics[width=0.45\textwidth]{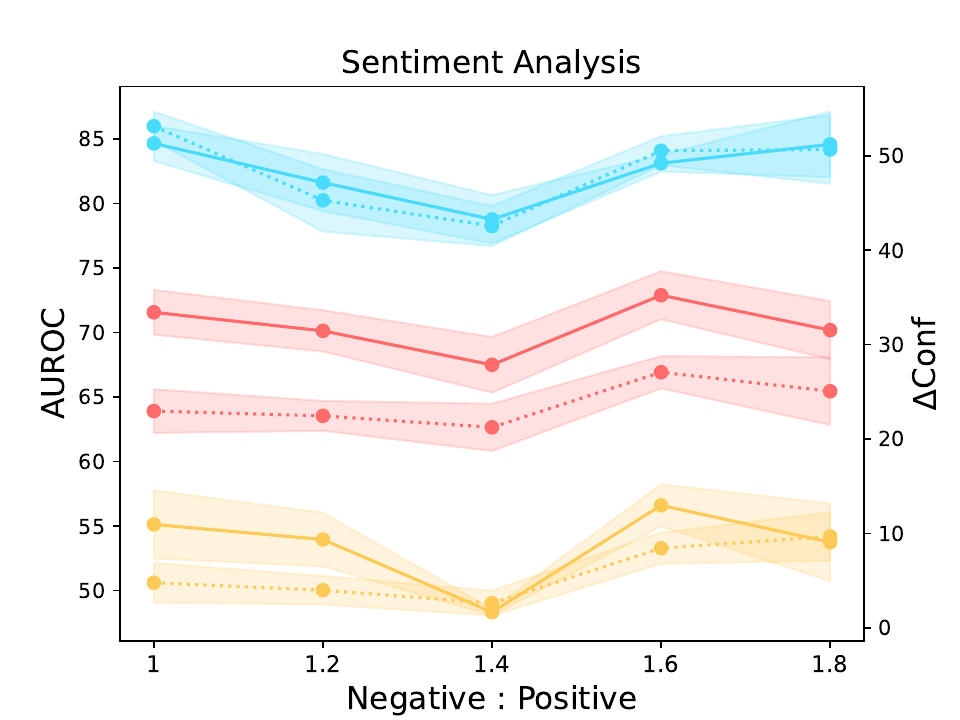}
%   \hspace{1in}
  \includegraphics[width=0.45\textwidth]{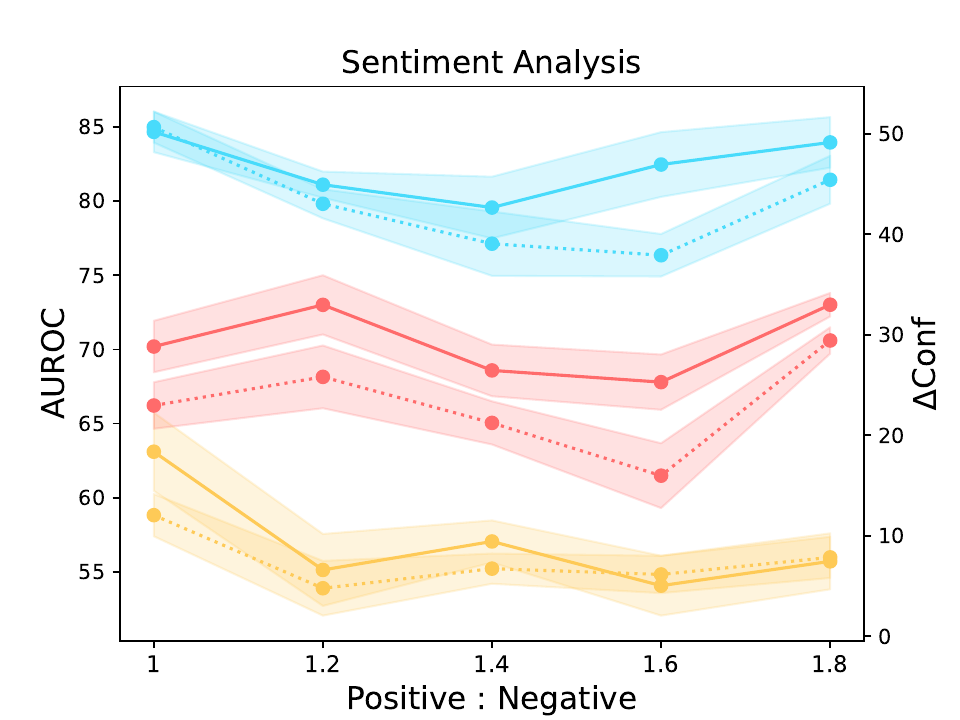}
%   \hspace{1in}
   \hspace{1in}
   \includegraphics[width=0.95\textwidth]{figs/legend.pdf}
  % \vspace{-10pt}
  \caption{Further analysis of the dataset imbalance issue. The evaluation datasets are listed in Table~\ref{tab:dataset}.}
   \label{fig:quantify_further}
\end{figure*}
% \vspace{-10pt}

%% file: tabs/ablation.tex
\begin{table}[]

\centering
\resizebox{0.49\textwidth}{!}{
\begin{tabular}{c|l|rrrrrr}
\toprule
\multirow{7}{*}{Amazon} & Dataset              & \multicolumn{2}{c}{Amazon}                                   & \multicolumn{2}{c}{SST-5}                                    & \multicolumn{2}{c}{SemEval}                                  \\ \cmidrule(l){2-8} 
                        & Method               & \multicolumn{1}{c}{AUROC} & \multicolumn{1}{c}{$\Delta$Conf} & \multicolumn{1}{c}{AUROC} & \multicolumn{1}{c}{$\Delta$Conf} & \multicolumn{1}{c}{AUROC} & \multicolumn{1}{c}{$\Delta$Conf} \\ \cmidrule(l){2-8} 
                        & LM-TOAST \includegraphics[width=0.017\textwidth]{figs/toast.png}             & \textbf{87.44 (1.12)}     & \textbf{42.46 (3.03)}            & \textbf{79.32 (1.09)}     & 26.37 (1.04)                     & 73.17 (2.36)              & 20.21 (1.70)                     \\
                        & w/o Cross-annotation & 87.35 (0.43)              & 27.72 (1.28)                     & 66.96 (0.58)              & 8.20 (1.84)                      & 72.41 (1.27)              & 8.11 (1.84)                      \\
                        & w/o Down-sampling    & 81.50 (3.32)              & 19.45 (2.67)                     & 77.03 (1.36)              & 20.46 (1.32)                     & 71.69 (0.13)              & 15.90 (0.70)                     \\
                        & w/o Augmentation     & 84.37 (2.00)              & 37.88 (2.89)                     & 74.14 (0.86)              & 19.73 (0.26)                     & 74.09 (2.24)     & 17.76 (0.35)                     \\
                        & w/o Decay $\alpha$            & 80.56 (4.13)              & 17.57 (4.13)                     & 77.92 (0.21)              & \textbf{29.34 (0.99)}            & \textbf{75.18 (0.23)}     & \textbf{22.10 (1.16)}            \\ \bottomrule
\end{tabular}
}
% \vspace{-pt}

\caption{The ablation study of various components in LM-TOAST.}
\label{tab:ablation}
\end{table}

%% file: figs/app1.tex
\begin{figure*}
  \centering
  \includegraphics[width=0.32\textwidth]{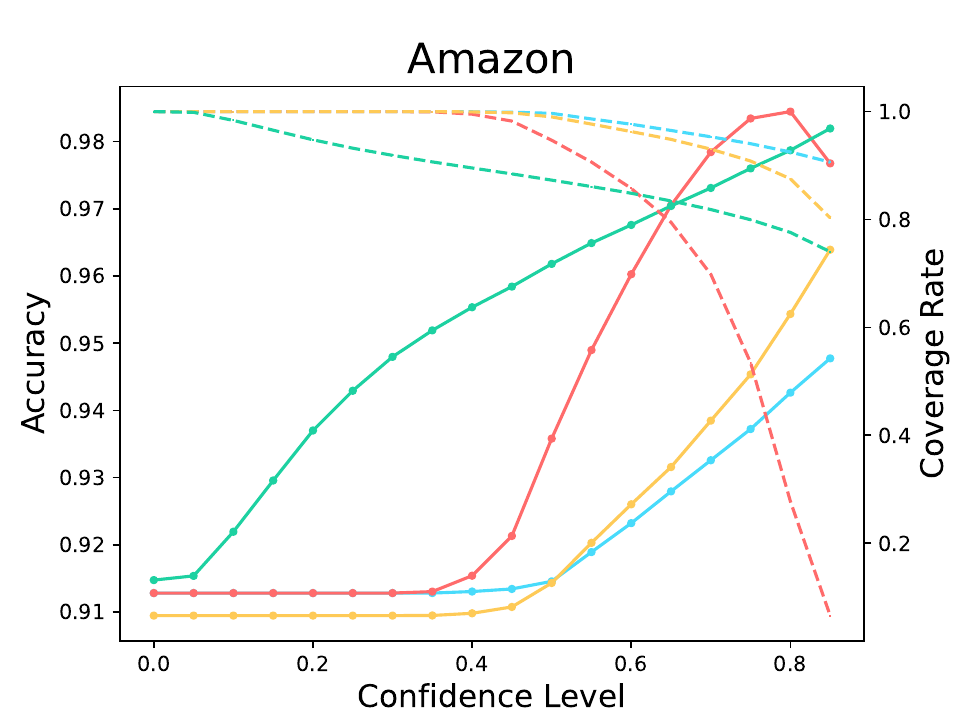}
%   \hspace{1in}
  \includegraphics[width=0.32\textwidth]{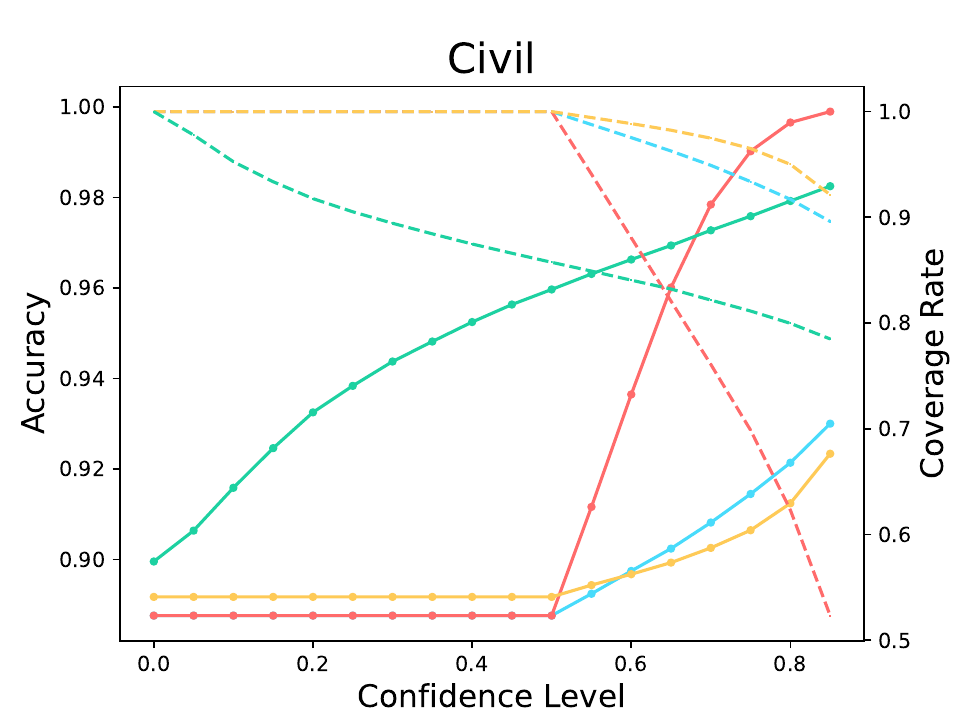}
%   \hspace{1in}
  \includegraphics[width=0.32\textwidth]{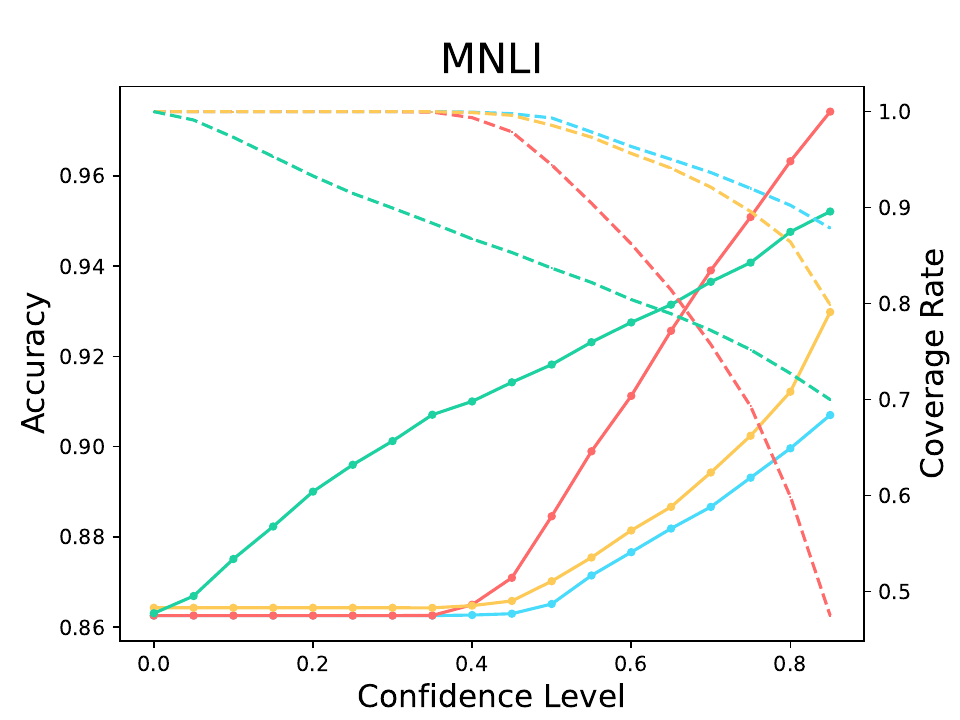}
   \hspace{1in}
   \includegraphics[width=0.75\textwidth]{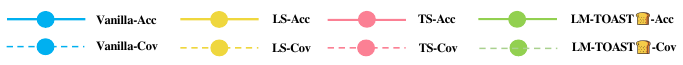}
  \vspace{-10pt}
  \caption{Results of selective classification. The -Acc and -Conv in the legend denote Accuracy and Coverage rate respectively.
  LM-TOAST steadily increases performance when the confidence level keeps getting larger.}
  
   \label{fig:app1}
\end{figure*}

%% file: figs/ablation.tex
\begin{figure}
  \centering
  \includegraphics[width=0.45\textwidth]{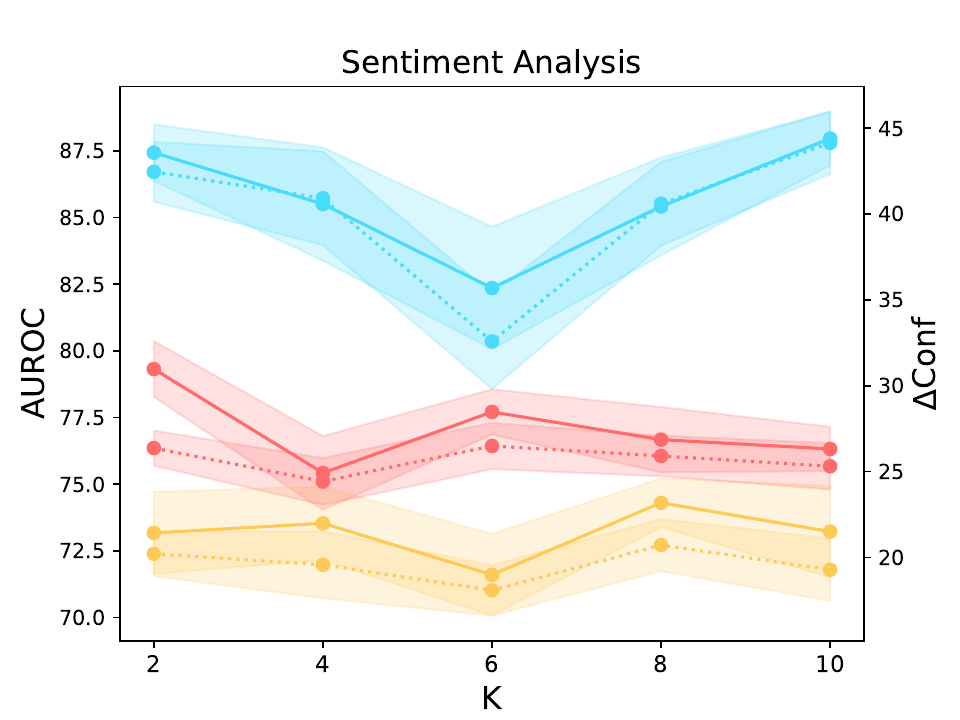}
%   \hspace{1in}
   \hspace{1in}
   \includegraphics[width=0.45\textwidth]{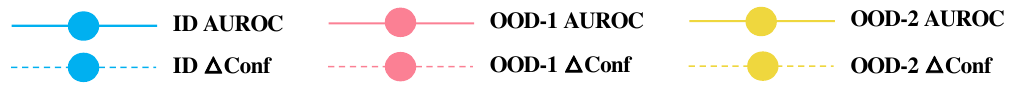}
  % \vspace{-10pt}
  \caption{The influence of K in the cross-annotation split process. The evaluation datasets are listed in Table~\ref{tab:dataset}.}
  \label{fig:ablation}
\end{figure}

%% file: tabs/app1.tex
\begin{table}[]

\centering
\resizebox{0.49\textwidth}{!}{
\begin{tabular}{l|l|cccccc}
\toprule
\multicolumn{1}{c|}{\multirow{6}{*}{Amazon}} & Dataset  & \multicolumn{2}{c}{Amazon}      & \multicolumn{2}{c}{SST-5}       & \multicolumn{2}{c}{SemEval}     \\ \cmidrule(l){2-8} 
\multicolumn{1}{c|}{}                        & Method   & AUROC$_{risk}$$\downarrow$          & Cov$\uparrow$            & AUROC$_{risk}$$\downarrow$          & Cov$\uparrow$            & AUROC$_{risk}$$\downarrow$          & Cov$\uparrow$            \\ \cmidrule(l){2-8} 
\multicolumn{1}{c|}{}                        & Vanilla  & 14.20          & 89.82          & 20.86          & 21.37          & 28.32          & 18.40          \\
\multicolumn{1}{c|}{}                        & TS       & 14.27          & 90.22          & 20.86          & 23.24          & 28.11          & 18.82          \\
\multicolumn{1}{c|}{}                        & LS       & 18.07          & 89.28          & 23.19          & 2.81           & 29.48          & 13.35          \\
\multicolumn{1}{c|}{}                        & LM-TOAST \includegraphics[width=0.017\textwidth]{figs/toast.png} & \textbf{12.56} & \textbf{91.27} & \textbf{20.68} & \textbf{23.52} & \textbf{26.83} & \textbf{24.25} \\ \midrule
\multirow{6}{*}{Civil}                       & Dataset  & \multicolumn{2}{c}{Civil}       & \multicolumn{2}{c}{Hate Speech} & \multicolumn{2}{c}{Implicit}        \\ \cmidrule(l){2-8} 
                                             & Method   & AUROC$_{risk}$$\downarrow$          & Cov$\uparrow$            & AUROC$_{risk}$$\downarrow$          & Cov$\uparrow$            & AUROC$_{risk}$$\downarrow$          & Cov$\uparrow$            \\ \cmidrule(l){2-8} 
                                             & Vanilla  & 9.67           & 84.71          & 37.20          & 10.46          & \textbf{34.01} & 0.04           \\
                                             & TS       & 9.67           & 84.71          & 37.20          & 10.46          & \textbf{34.01} & \textbf{0.16}  \\
                                             & LS       & 8.85           & 85.17          & 37.83          & 1.05           & 36.08          & 0              \\
                                             & LM-TOAST \includegraphics[width=0.017\textwidth]{figs/toast.png} & \textbf{7.99}  & \textbf{88.03} & \textbf{34.45} & \textbf{18.83} & \textbf{34.01} & \textbf{0.16}  \\ \midrule
\multirow{6}{*}{MNLI}                        & Dataset  & \multicolumn{2}{c}{MNLI}        & \multicolumn{2}{c}{HANS}        & \multicolumn{2}{c}{ANLI}        \\ \cmidrule(l){2-8} 
                                             & Method   & AUROC$_{risk}$$\downarrow$          & Cov$\uparrow$            & AUROC$_{risk}$$\downarrow$          & Cov$\uparrow$            & AUROC$_{risk}$$\downarrow$          & Cov$\uparrow$            \\ \cmidrule(l){2-8} 
                                             & Vanilla  & 17.40          & 69.63          & 47.04          & 62.45          & \textbf{55.89} & -              \\
                                             & TS       & 17.40          & 69.75          & 47.04          & 0              & \textbf{55.89} & -              \\
                                             & LS       & 19.49          & 68.18          & 53.48          & 0              & 56.93          & \textbf{-}     \\
                                             & LM-TOAST \includegraphics[width=0.017\textwidth]{figs/toast.png} & \textbf{17.26} & \textbf{71.74} & \textbf{39.40} & \textbf{79.46} & 56.03          & -              \\ \bottomrule
\end{tabular}
}
% \vspace{-pt}

\caption{Experimental results of selective classification. $\downarrow$: Lower is better. $\uparrow$: Higher is better. Results in ANLI are left out due to PLMs' low performance.}
\label{tab:app1}
\end{table}

%% file: tabs/prompt.tex
\begin{table*}[]
\centering
\resizebox{0.85\textwidth}{!}{
\begin{tabular}{l|l|c|c}
\toprule
Task                                                                                      & Dataset     & \multicolumn{1}{l|}{Template}                                                                                                                                                                                       & Verbalizer               \\ \midrule
\multirow{3}{*}{\begin{tabular}[c]{@{}l@{}}\\ Sentiment \\ Analysis\end{tabular}}            & Amazon      & It was \textless{}mask\textgreater{}. \{”placeholder”: ”text a”\}                                                                                                                                                   & {[}bad, good, neutral{]} \\ \cmidrule(l){2-4} 
                                                                                          & SST-5       & It was \textless{}mask\textgreater{}. \{”placeholder”: ”text a”\}                                                                                                                                                   & {[}bad, good, neutral{]} \\ \cmidrule(l){2-4} 
                                                                                          & SemEval     & It was \textless{}mask\textgreater{}. \{”placeholder”: ”text a”\}                                                                                                                                                   & {[}bad, good, neutral{]} \\ \midrule
\multirow{3}{*}{\begin{tabular}[c]{@{}l@{}}  Hate \\ Speech \\ Detection\end{tabular}}      & Civil       & It was \textless{}mask\textgreater{}. \{”placeholder”: ”text a”\}                                                                                                                                                   & {[}benign, toxic{]}      \\ \cmidrule(l){2-4} 
                                                                                          & Hate Speech & It was \textless{}mask\textgreater{}. \{”placeholder”: ”text a”\}                                                                                                                                                   & {[}benign, toxic{]}      \\ \cmidrule(l){2-4} 
                                                                                          & Implicit    & It was \textless{}mask\textgreater{}. \{”placeholder”: ”text a”\}                                                                                                                                                   & {[}benign, toxic{]}      \\ \midrule
\multirow{3}{*}{\begin{tabular}[c]{@{}l@{}}\\ \\ \\ Natural \\ Language \\ Inference\end{tabular}} & MNLI        & \begin{tabular}[c]{@{}c@{}}Given the two sentences:\\ (1) \{”placeholder”: ”text a”\}.\\ (2) \{”placeholder”: ”text b”\}.\\ Does the first sentence entails the second ? \textless{}mask\textgreater{}\end{tabular} & {[}No, Yes, Maybe{]}     \\ \cmidrule(l){2-4} 
                                                                                          & HANS        & \begin{tabular}[c]{@{}c@{}}Given the two sentences:\\ (1) \{”placeholder”: ”text a”\}.\\ (2) \{”placeholder”: ”text b”\}.\\ Does the first sentence entails the second ? \textless{}mask\textgreater{}\end{tabular} & {[}No, Yes, Maybe{]}     \\ \cmidrule(l){2-4} 
                                                                                          & ANLI        & \begin{tabular}[c]{@{}c@{}}Given the two sentences:\\ (1) \{”placeholder”: ”text a”\}.\\ (2) \{”placeholder”: ”text b”\}.\\ Does the first sentence entails the second ? \textless{}mask\textgreater{}\end{tabular} & {[}No, Yes, Maybe{]}     \\ \bottomrule
\end{tabular}
}
\caption{The manual templates and verbalizers adopted for the main task.}
\label{tab:template}
\end{table*}

%% file: tabs/template2.tex
\begin{table*}[]
\centering
\resizebox{0.85\textwidth}{!}{
\begin{tabular}{l|l|c|c}
\toprule
Task                                                                                      & Dataset     & \multicolumn{1}{l|}{Template}                                                                                                                                                                                                            & Verbalizer                         \\ \midrule
\multirow{3}{*}{\begin{tabular}[c]{@{}l@{}}\\ \\ Sentiment \\ Analysis\end{tabular}}            & Amazon      & \begin{tabular}[c]{@{}c@{}}Sentence: \{”placeholder”: ”text a”\} The predicted sentiment is \{”placeholder”: ”text b”\} .\\ Is the prediction True or False ? It’s \{”mask”\} .\end{tabular}                                             & \multirow{20}{*}{{[}False, True{]}} \\ \cmidrule(l){2-3} 
                                                                                          & SST-5       & \begin{tabular}[c]{@{}c@{}}Sentence: \{”placeholder”: ”text a”\} The predicted sentiment is \{”placeholder”: ”text b”\} .\\ Is the prediction True or False ? It’s \{”mask”\} .\end{tabular}                                             &                                    \\ \cmidrule(l){2-3} 
                                                                                          & SemEval     & \begin{tabular}[c]{@{}c@{}}Sentence: \{”placeholder”: ”text a”\} The predicted sentiment is \{”placeholder”: ”text b”\} .\\ Is the prediction True or False ? It’s \{”mask”\} .\end{tabular}                                             &                                    \\ \cmidrule(l){1-3} 
\multirow{3}{*}{\begin{tabular}[c]{@{}l@{}}\\ Hate \\  Speech \\ Detection\end{tabular}}   & Civil       & \begin{tabular}[c]{@{}c@{}}Sentence: \{”placeholder”: ”text a”\} The predicted toxicity is \{”placeholder”: ”text b”\} .\\ Is the prediction True or False ? It’s \{”mask”\} .\end{tabular}                                              &                                    \\ \cmidrule(l){2-3} 
                                                                                          & Hate Speech & \begin{tabular}[c]{@{}c@{}}Sentence: \{”placeholder”: ”text a”\} The predicted toxicity is \{”placeholder”: ”text b”\} .\\ Is the prediction True or False ? It’s \{”mask”\} .\end{tabular}                                              &                                    \\ \cmidrule(l){2-3} 
                                                                                          & Implicit    & \begin{tabular}[c]{@{}c@{}}Sentence: \{”placeholder”: ”text a”\} The predicted toxicity is \{”placeholder”: ”text b”\} .\\ Is the prediction True or False ? It’s \{”mask”\} .\end{tabular}                                              &                                    \\ \cmidrule(l){1-3} 
\multirow{3}{*}{\begin{tabular}[c]{@{}l@{}}\\ \\ Natural \\ Language \\ Inference\end{tabular}} & MNLI        & \begin{tabular}[c]{@{}c@{}}Given the two sentences: \{”placeholder”: ”text a”\}\\ The predicted relationship between the two sentences is \{”placeholder”: ”text b”\}\\ Is the prediction True or False ? It’s \{”mask”\} .\end{tabular} &                                    \\ \cmidrule(l){2-3} 
                                                                                          & HANS        & \begin{tabular}[c]{@{}c@{}}Given the two sentences: \{”placeholder”: ”text a”\}\\ The predicted relationship between the two sentences is \{”placeholder”: ”text b”\}\\ Is the prediction True or False ? It’s \{”mask”\} .\end{tabular} &                                    \\ \cmidrule(l){2-3} 
                                                                                          & ANLI        & \begin{tabular}[c]{@{}c@{}}Given the two sentences: \{”placeholder”: ”text a”\}\\ The predicted relationship between the two sentences is \{”placeholder”: ”text b”\}\\ Is the prediction True or False ? It’s \{”mask”\} .\end{tabular} &                                    \\ \bottomrule
\end{tabular}
}
\caption{The manual templates and verbalizers adopted for the calibration task.}
\label{tab:template2}
\end{table*}